\documentclass{article}

\usepackage{subfig}
\usepackage{cancel}
\usepackage{graphicx}
\usepackage[dvipsnames]{xcolor}
\usepackage{tabularx}
\usepackage{amsmath,amssymb}
\usepackage{amsthm}
\usepackage{algorithm}%
\usepackage{algorithmicx}%
\usepackage{algpseudocode}%
\usepackage{cases}
\usepackage{caption} 
\usepackage{gensymb}
\usepackage[left=2cm, right=2cm, top=2.5cm, bottom=2.5cm]{geometry}
\usepackage{gensymb} 
\usepackage{eufrak}
\usepackage{bm} 
\usepackage{hyperref}
\usepackage{cleveref}
\usepackage{mathtools} 
\usepackage{mathrsfs} 
\usepackage{marginnote}
\usepackage{multirow} 
\usepackage{multicol} 
\usepackage{numprint} 
\usepackage{longtable}
\usepackage{lineno} 
\usepackage{ulem}
\usepackage{siunitx}
\usepackage{rotating}
\usepackage{titlesec}
\usepackage{upgreek}

\newcommand{\partialder}[2]{\partial_{#2}#1}
\newcommand{\der}[2]{\frac{d#1}{d#2}}
\newcommand{\secondpartialder}[2]{\frac{\partial^2#1}{\partial#2^2}}
\newcommand{\partialderlong}[2]{ \frac{\partial}{\partial#2}\left(#1\right) }

\newcommand{\loss}[0]{\mathcal{L}}

\newcommand{\E}[0]{\mathbb{E}}
\newcommand{\db}[0]{\Delta b}
\newcommand{\RR}[0]{\mathbb{R}}

\newtheorem{theorem}{Theorem}[section]

\theoremstyle{definition}              

\newtheorem{remark}[theorem]{Remark}

\begin{document}
\title{\textbf{Elimination-compensation pruning for fully-connected neural networks}}
\author{
Enrico Ballini$^1$\thanks{These authors contributed equally to this work.} \and
Luca Muscarnera$^1$\footnotemark[1] \and
Alessio Fumagalli$^1$ \and
Anna Scotti$^1$ \and
Francesco Regazzoni$^1$\thanks{Corresponding author: \texttt{francesco.regazzoni@polimi.it}} 
\\[0.5em]
\textit{$^1$MOX, Department of Mathematics, Politecnico di Milano,}\\
\textit{Piazza Leonardo da Vinci 32, 20133 Milano, Italy}
}
\maketitle


\begin{abstract}

The unmatched ability of Deep Neural Networks in capturing complex patterns in large and noisy datasets is often associated with their large hypothesis space, and consequently to the vast amount of parameters that characterize model architectures. Pruning techniques affirmed themselves as valid tools to extract sparse representations of neural networks parameters, carefully balancing  between compression and preservation of information. However, a fundamental assumption behind pruning is that expendable weights should have small impact on the error of the network, while highly important weights should tend to have a larger influence on the inference. 
We argue that this idea could be generalized; what if a weight is not simply removed but also compensated with a perturbation of the adjacent bias, which does not contribute to the network sparsity? Our work introduces a novel pruning method in which the importance measure of each weight is computed considering the output behavior after an optimal perturbation of its adjacent bias, efficiently computable by automatic differentiation. These perturbations can be then applied directly after the removal of each weight, independently of each other. After deriving analytical expressions for the aforementioned quantities, numerical experiments are conducted to benchmark this technique against some of the most popular pruning strategies, demonstrating an intrinsic efficiency of the proposed approach in very diverse machine learning scenarios. 
Finally, our findings are discussed and the theoretical implications of our results are presented.

\end{abstract}

\section{Introduction}

Nowadays, neural networks are capable of succeeding in very complex tasks spanning different applications such as, to mention a few, computer vision \cite{Voulodimos2018}, natural language processing \cite{Otter2021}, scientific computing and numerical solution of partial differential equations \cite{EdDyyany2025, Raissi2019, Lu2021}.  
Neural networks rely on a potentially very large number of parameters that must be optimized for the specific task \cite{Raissi2019, Karlbauer2021, Ballini2025}. Such a high parameter count can give rise to challenges related to training efficiency, storing memory, and inference time, especially when considering the possible embedding of neural networks in portable devices with limited hardware resources.  
Despite the excellent ability of neural networks to solve complex problems, their application can be limited by the computational burden arising from the sequence of matrix-vector multiplications required to perform inference,  
thus restricting their use in several domains where computational resources are constrained by structural or energy limitations.  
Moreover, it is interesting to investigate whether the large number of parameters is truly necessary to achieve good performance.

It is therefore relevant to study methodologies for handling the large amount of parameters issue.  
There can be different strategies: to mention a few, it is possible to compress the weight matrices with a proper factorization; reduce numerical precision, for instance by using 8-bit representations instead of 64-bit ones; restrict parameters to a finite set of values via quantization; apply knowledge distillation; study better neural network architectures; or merge neurons with highly correlated activations \cite{Babaeizadeh2016}.  
We refer, among many review papers, to \cite{Alqahtani2021, Pant2021, Vadera2022, Cheng2024} for a deep review of the aforementioned methods, and to \cite{Gale2019} for a comparison of well-established techniques.

In this work, we focus on pruning methods, which are a family of techniques in which some non-important synapses are removed (the weights are set to zero), leading to a lower number of non-zero weights.  
The authors of \cite{Qian2021} explain that it is highly probable that there exists a pruned network close enough to the original one, which motivates the study of pruning method.  

The core idea is to first define an \textit{importance} measure of the weight then, starting from a possibly already trained large neural network, set to zero as many weights as possible whose importance measure is low, while preserving the accuracy of the network.

Several techniques were developed in this direction and we refer to \cite{Reed1993, Blalock2020, Cheng2024} for an overview. \\
Pruning is typically executed in a finite loop of train, then prune, then repeat, which usually reduces to train-prune-train, where the last train phase is referred to as \textit{fine train}.  
It is important to mention that pruning can be an effective strategy to facilitate training and obtain a good architecture for the specific problem \textit{a priori} to training \cite{Ballini2025phd}.  
This enhancement can be achieved by incorporating information about the specific problem at hand into the architecture of the network.  
In our work, we focus on a pruning strategy for generic tasks.
Pruning can be achieved by a \textit{local} technique or a \textit{global} technique. In the former, the weights are set to zero by considering local quantities, such as their magnitude \cite{Janowsky1989}, or more complex importance indicators that are computed by considering the quantities of the isolated layer \cite{Engelbrecht1996, Zurada1997, Mauch2018}.\\
In the latter, the importance of the weight is defined by considering quantities related to the output of the neural network. A well-known strategy is Optimal Brain Damage \cite{LeCun1989}, Optimal Brain Surgeon \cite{Hassibi1992}, and their variants, which compute the second-order derivatives of the loss and remove weights with low influence on it.

Other strategies focus on sensitivity, considering the first-order derivative of the loss and, possibly, derived quantities such as variance and mean values. One of the first works in this direction is \cite{Mozer1988}, while more recent works include \cite{Xiao2019, Molchanov2019, Sanh2020}. 
It is also possible to effectively avoid the use of derivatives, as shown in \cite{Yu2018}, obtaining a global measurement of the weight importance. \\
In contrast to global approach, \cite{Engelbrecht2001} considers statistical quantities of the sensitivity, a line of research that is further developed by \cite{Fnaiech2004}, who make the computation of importance dependent on local quantities. \\
We focus on a global approach, which has the advantage of being more suitable for formal analysis.  

From the cited literature, it is apparent that the derivative of the loss function w.r.t. the trainable weights and the value of the weights are relevant quantities for obtaining effective pruning. Indeed, \cite{Blalock2020} proposes the product of these quantities as a possible importance measure.  \\
To the extent of the author's knowledge, little has been done regarding the inclusion of biases in pruning methods, although biases can be relevant in defining importance \cite{Mauch2017, Mauch2018}.

\paragraph{Contribution and paper organization.}
In this paper, we propose a pruning technique for fully connected neural networks applied to different tasks with emphasis on applications in scientific machine learning. Our method is based on the minimization of the expected value of the discrepancy between the output of the original network and the pruned one.  
Remarkably, we never compute the derivatives of the loss function which can be subject to noise effects nearby a minimum value; instead, using a Taylor expansion of the network, we define a novel importance measure that concurrently includes the effects of both the weights on the network output and the effects of the biases. \\
The paper is organized as follows. In \Cref{sec:basic}, we present a general framework for pruning. In \Cref{sec:method}, we introduce our proposed method. Numerical test cases are presented in \Cref{sec:numerical_cases}. Conclusions are drawn in \Cref{sec:conclusion}.


\section{Basic principles}\label{sec:basic}
We present in this section the basic definitions and notation for neural networks, \Cref{sec:neural_network}, and pruning in general form, \Cref{sec:general_pruning}.

\subsection{Fully-connected neural networks}\label{sec:neural_network}
The goal of this section is setting some notation regarding fully-connected neural networks.  Considering a layer $\ell \in 1, \ldots, L$, whose size is denoted by $n_\ell$, we have the following relation between the input, $z^{\ell-1} \in \mathbb{R}^{n_\ell-1}$, and the output, $z^{\ell}\in\mathbb{R}^{n_\ell}$, of the layer $\ell$:
\begin{align*}
	z^\ell = \sigma^\ell(W^{\ell} z^{(\ell-1)} + b^{\ell}),
\end{align*}
where the coefficients $W^{\ell}\in\mathbb{R}^{n_\ell\times n_{\ell-1}}$ and $b^{\ell}\in\mathbb{R}^{n_\ell}$ are the trainable weights of the unknown affine transformation of layer $\ell$, and $\sigma^{\ell}$ is the non-linear activation function, except for $\sigma^L$ which is set to be the identity function.  
We call the term $W_{ij}^\ell z_j^{\ell-1}$ the \textit{signal} of the layer $\ell$.
With abuse of notation, calling $z^{\ell}(z^{(\ell-1)}) \coloneq \sigma^{\ell}(W^{\ell}z^{(\ell-1)} + b^{\ell})$, the full neural network can be written as
\begin{equation*}
	y  = z^{(L)} \circ z^{(L-1)}  \circ \ldots \circ z^{(1)} (x),
\end{equation*}
with $x \in \mathbb{R}^{n_\text{in}}$ the input and $y \in \mathbb{R}^{n_\text{out}}$ the output. To simplify the notation, we define $\theta = \cup_\ell(W^{\ell}, b^{\ell})$ to be all the trainable weights of a neural network and we denote by $|\theta|$ the total number of trainable weights.
When it is relevant to highlight the dependence of the network on its trainable weights, we use the following notation 
\begin{equation*}
    y = y(x; \theta).   
\end{equation*}
The value of the parameters, $\theta$, are optimized to minimize a loss function, $\loss$.

\subsection{Why is pruning complicated? A formal approach}\label{sec:general_pruning}

For clarity of exposition, we introduce the \textit{mask matrix}, denoted by $M_{ij}^\ell$, associated with layer $\ell$ and taking values in $\mathcal{M}^\ell = \{ M \in \RR^{n_\ell \times n_{\ell-1}} : M_{ij} \in \{0,1\} \}$. Its entries indicate which weights are kept during a training procedure and which are set to zero, pruning the corresponding connections. The pruned layer is given by
\begin{align*}
    z^\ell = \sigma^\ell(M^\ell\odot W^{\ell} z^{(\ell-1)} + b^{\ell}).
\end{align*}
We denote by $\overline{\theta}_M$ the set of the weights with the masks applied. Therefore, the pruned neural network will be denoted by $y(x, \overline{\theta}_M)$.
We also write $|M^\ell|$ to denote the number of nonzero entries of $M^\ell$.

The pruning problem can be formulated within the following general framework\footnote{We do not use Einstein summation convention.}:
\begin{align}\label{eq:general_opt}
    \begin{aligned}
        &\min_{\substack{M_{ij}^{\ell}\in\{0,1\} \\ \forall i,j,\ell}}  
        \rho\left(y(x; \theta),\, y(x; \overline{\theta}_{M_{ij}^\ell})\right), \\
        &\text{subject to:} \\
        &\sum_\ell |M^{\ell}| = N,
    \end{aligned}
\end{align}
where $\rho$ denotes an appropriate measure of the discrepancy between the original network and its pruned counterpart.

Given the extremely large number of parameters involved, the optimization problem \eqref{eq:general_opt} is computationally infeasible. It is therefore commonly replaced by the following surrogate procedure. To determine which weights should be removed, we assign to each weight $W_{ij}^\ell$ an importance value $\mathcal{I}_{W_{ij}^\ell}$. Weights whose importance falls below a prescribed threshold are then set to zero:
\begin{align}
    \begin{aligned}
        &\mathcal{I}_{W_{ij}^\ell} = 
        d\left(y(x; \theta), y\left(x; \left.\theta\right|_{W_{ij}^\ell\leftarrow 0}\right)\right), \\
        &M_{ij}^\ell = 0,\ \ \text{for } i,j,\ell \ \ \text{s.t.}\ \ \mathcal{I}_{W_{ij}^\ell} \leq T_h .
    \end{aligned}
\end{align}
Here $T_h$ is a threshold that may be chosen a priori or determined a posteriori via order statistics to obtain a prescribed number of retained weights. As in \eqref{eq:general_opt}, the function $d$ serves as a suitable measure of discrepancy between the original and pruned networks. Examples of choices of importance scores are
$\mathcal{I}_{W_{ij}^\ell} = \frac{1}{2}\secondpartialder{\loss}{{W_{ij}^\ell}}(W_{ij}^\ell)^2$ \cite{LeCun1989}, 
$\mathcal{I}_{W_{ij}^\ell} = |W_{ij}^\ell|$ \cite{Janowsky1989, Blalock2020}, or possibly 
$\mathcal{I}_{W_{ij}^\ell} = \left|\loss(y(x; \theta))-\loss(y(x; \left.\theta\right|_{W_{ij}^\ell\leftarrow0}))\right|$.

\begin{remark}[Pruning for inefficient training]
Let $\mathcal{Y}$ be the set of neural networks associated with a given architecture. The training procedure aims to identify the optimal (in the sense specified by the loss function) neural network in $\mathcal{Y}$. We remark that the pruned neural network is included in $\mathcal{Y}$. Consequently, pruning does not confer any clear theoretical advantage. Its value lies instead in practical considerations: training can be a challenging task, and the large number of parameters to be optimized may prevent the procedure from finding a suitable candidate for minimizing the loss function. Pruning seeks to simplify the optimization problem while preserving the essential characteristics of the network.
\end{remark}

\section{Elimination-compensation approach}\label{sec:method}
In this section, we present our proposed method. The central idea is to compensate the removal of a weight by modifying the corresponding bias, and to incorporate this adjustment into the definition of the importance measure. In essence, the importance quantifies how strongly a given weight influences the network output and how well it can be replaced by an appropriate bias correction.

Therefore, rather than simply setting a weight to zero, we compensate for its removal by adjusting the bias $b_i^\ell$. Denoting the adjustment by $\db^\ell$, the $\ell$-th layer of the pruned network becomes
\begin{align*}
    z^\ell = \sigma^\ell(M^\ell\odot W^{\ell} z^{(\ell-1)} + b^{\ell} + \db^\ell).
\end{align*}
We highlight that the modification of the bias does not affect the inference cost of the network, as $b^\ell + \db^\ell$ will be replaced by the result of the summation.

For clarity of exposition, we illustrate the main steps for the scalar output case, $n_\text{out} = 1$, and provide the general formulation for $n_\text{out} \geq 1$ thereafter.

Before giving a formal definition of the importance measure, we introduce the discrepancy $\Delta y$ between the output of the pruned network and that of the original one:
\begin{align*}
    \Delta y(x; \theta, i,j,\ell, \db_{ij}^\ell) = 
    y(x; \theta) -
    y\left(x; \left.\theta\right|_{W_{ij}^\ell\leftarrow0, b_i^\ell\leftarrow b_i^\ell + \sum_j\db_{ij}^\ell}\right),
\end{align*}
where $\db_{ij}^\ell$\footnote{Note the presence of double subscript, $ij$, in the compensation $\db_{ij}^\ell$ as it is associated to each weight $W_{ij}^\ell$} is a bias compensation term that is chosen appropriately. 

\paragraph{Direct non-linear method.}
We define the importance of a weight as the expected output variation over the train dataset obtained when the weight is set to zero and compensated by an appropriate value $\db_{ij}^\ell$:
\begin{align*}
    \mathcal{I}_{W_{ij}^\ell}
    = \E_x\!\left[\, \Delta y(x; \theta, i,j,\ell, \db_{ij}^\ell)^2 \right].
\end{align*}
It remains to determine an optimal choice for $\db_{ij}^\ell$.

A first idea is to choose $\db_{ij}^\ell$ so as to preserve the mean value of the signal (\Cref{sec:neural_network}):
\begin{align*}
    \overline{\db_{ij}^\ell}
    \coloneq \E_x\!\left[ W_{ij}^\ell z_j^{\ell-1}\right]
    = W_{ij}^\ell\, \E_x\!\left[z_j^{\ell-1}\right].
\end{align*}
The corresponding importance then becomes
\begin{align}\label{eq:imp_nonlinear}
    \mathcal{I}_{W_{ij}^\ell}
    = \E_x \!\left[\, \Delta y(x; \theta, i,j,\ell, \overline{\db_{ij}^\ell})^2 \right].
\end{align}
This simple and direct approach encounters a practical limitation: the importance must be computed on a weight-by-weight basis, meaning that at each step a single weight is set to zero while all others remain unchanged. This would require a number of forward evaluations equal to the number of weights, rendering the procedure computationally prohibitive.
Indeed, denoting by $C_f$ the cost of the forward evaluation, measured as the number of floating-point operations, the computation of the importance of all the weights involves a cost that depends on the total number of weights proportional to $|W|C_f$.
Due to the high computational cost, we must adopt an alternative strategy.

\paragraph{Proposed method.}
Assuming that the weights and bias adjustments are small, we proceed by computing a Taylor expansion of the network to approximate the discrepancy:
\begin{align}\label{eq:discrepancy_linearized}
\begin{aligned}
    \Delta y(x; \theta, i,j,\ell, \db_{ij}^\ell) \approx \delta y(x; \theta, i,j,\ell, \db_{ij}^\ell) &\coloneq y(x;\theta) - \left( y(x; \theta) + \partialder{y}{W_{ij}^  \ell}(x; \theta)(0-W_{ij}^\ell) + \partialder{y}{b_i^\ell}(x; \theta)\db_{ij}^\ell \right) = \\
    &= \partialder{y}{W_{ij}^\ell}\:W_{ij}^\ell - \partialder{y}{b_i^\ell}\:\db_{ij}^\ell.
\end{aligned}
\end{align}
We can now define the importance by using the linear approximation (Eq.~\eqref{eq:discrepancy_linearized}) and determining the optimal compensation through an associated minimization problem:

\begin{align}\label{eq:importance_comp_1}
    \mathcal{I}_{W_{ij}^\ell} = \min_{\db_{ij}^\ell \in \mathbb{R}}\E_x[\delta y(x; \theta, i,j,\ell, \db_{ij}^\ell)^2].
\end{align}
This optimization problem, although seemingly complicated, can be solved explicitly to obtain the optimal compensation $\widetilde{\db_{ij}^\ell}$. To identify the minimizer, we compute the following derivative:
\begin{align}
   &\partialderlong{\E_x[\delta y(x; \theta, i,j,\ell, \db_{ij}^\ell)^2]}{\db_{ij}^\ell} = \E_x\left[\left(\partialder{y}{W_{ij}^\ell}\:W_{ij}^\ell - \partialder{y}{b_i^\ell}\:\db_{ij}^\ell\right)(-\partialder{y}{b_i^\ell})\right],
\end{align}
from which, by setting the derivative to zero, we obtain the optimal compensation $\db_{ij}^\ell = \widetilde{\db_{ij}^\ell}$:

\begin{equation}
    \widetilde{\db_{ij}^\ell} = \frac{W_{ij}^\ell \E_x \left[
         \partialder{y}{W_{ij}^\ell}\: \partialder{y}{b_i^\ell}\right]}{
         \E_x \left[
         \partialder{y}{b_i^\ell}\right]^2
         }.
\end{equation}
Replacing the expression of $\widetilde{\db_{ij}^\ell}$ into \eqref{eq:importance_comp_1} we have:
\begin{align*}
     \mathcal{I}_{W_{ij}^\ell}  = \E_x\left[\left( \partialder{y}{W_{ij}^\ell}\:W_{ij}^\ell - \frac{W_{ij}^\ell \E_x \left( \partialder{y}{W_{ij}^\ell}\: \partialder{y}{b_i^\ell}\right)}{\E_x \left(\partialder{y}{b_i^\ell}\right)^2}\partialder{y}{b_i^\ell}\right)^2\right].
\end{align*}
A note must be done on the computational cost: considering $\alpha C_f$ as the cost of backpropagation, the computation of the importance for all weights is proportional to $\alpha C_f$ instead of $|W|C_f$ as previously mentioned, making this strategy computationally affordable.

In the case of vectorial output, using the $\ell_2$-norm of the discrepancy to measure the error, we obtain the following expression:
\begin{align*}
    \mathcal{I}_{W_{ij}^\ell} = \E_x \left[ \sum_k \left( \partialder{y_k}{W_{ij}^\ell}\:W_{ij}^\ell -  \frac{\sum_k W_{ij}^\ell\E_x(\partialder{y_k}{W_{ij}^\ell}\:\partialder{y_k}{b_i^\ell})}{\sum_k\E_x[(\partialder{y_k}{b_i^\ell})^2]}\partialder{y_k}{b_i^\ell} \right)^2 \right].
\end{align*}

\section{Numerical Experiments}\label{sec:numerical_cases}

In this section, we present numerical experiments to demonstrate the effectiveness of the proposed method.
In the first experiment, we use the well-known MNIST dataset \cite{LeCun1998}, while in the second we employ a dataset constructed from solutions of partial differential equations \cite{Takamoto2022}. \\
The datasets are split into training and test sets. The training of the networks and the evaluation of importance are carried out using the training set. Subsequently, performance is assessed on the test set.
Performance is evaluated by computing the test loss for different pruning ratios. We define the pruning ratio, $r$, as the fraction of removed weights, namely
\begin{align*}
	r = 1 - \frac{|W|}{|W_\text{original}|},
\end{align*}
where $|W|$ denotes the total number of weights of the considered network\footnote{We do not explicitly refer to pruned weights, as the above definition will also apply to fully connected networks.}, summed over all layers, and $W_\text{original}$ is the total number of weights before pruning.
For clarity, we will report only the test losses. \\
The proposed method is compared against $5$ different strategies, listed below:
\begin{itemize}
	\item ``Non-linear'': This is the brute-force approach, in which importance is defined in \eqref{eq:imp_nonlinear}. This strategy is computationally expensive as its cost depends linearly on $|W|$, see \Cref{sec:method}, and therefore not practical. Nonetheless, for the sake of completeness, it is included, with the understanding that it is often infeasible for real applications.
	\item ``Magnitude'': A simple yet effective method \cite{Blalock2020}. In this strategy, importance is defined as $\mathcal{I}_{W_{ij}^{\ell}} = |W_{ij}^{\ell}|$.
	\item ``Gradient-Magnitude'': Here, importance is defined as $\mathcal{I}_{W_{ij}^{\ell}} = |W_{ij}^{\ell}|\der{\loss}{W_{ij}^\ell}$ \cite{Blalock2020}.
	\item ``Random'': In this strategy, the weights set to zero are randomly selected. Comparing with this strategy is relevant to demonstrate that the proposed method captures meaningful information during pruning. Nevertheless, random pruning can occasionally yield reasonable results \cite{Gale2019}.
	\item ``Fully-connected'': No pruning is applied. Instead, the number of weights is reduced by uniformly decreasing the width of the layers. The pruning strategy is considered effective if the pruned networks outperform an equivalent fully connected network, i.e. one with the same number of non-zero weights.
\end{itemize}

All experiments are repeated for $5$ different weight initializations to ensure the reliability of the results.

Training is performed using the well-known Adam optimizer \cite{Kingma2014} with default hyperparameters.

\subsection{Case 1: Classification}
In this case, we use the MNIST dataset \cite{LeCun1998, mnistweb}. The pruning strategies are applied to two different architectures, summarized in Tab.~\ref{tab:architectures_mnist}, to ensure the reliability of the results. The PReLU activation function is adopted in each layer. \\
Since a classification task is considered, the cross-entropy is used as loss function. 
The networks are initially trained for $15$ epochs, at which point pruning is applied. This is followed by an additional $15$ epochs of fine-tuning. The procedure can thus be described as train-prune-train.

\begin{figure}[h]
    \centering
    \begin{minipage}{0.9\textwidth}
    \centering\textbf{Architecture 1} \\
    \vspace{1mm}
    \includegraphics[width=0.48\linewidth]{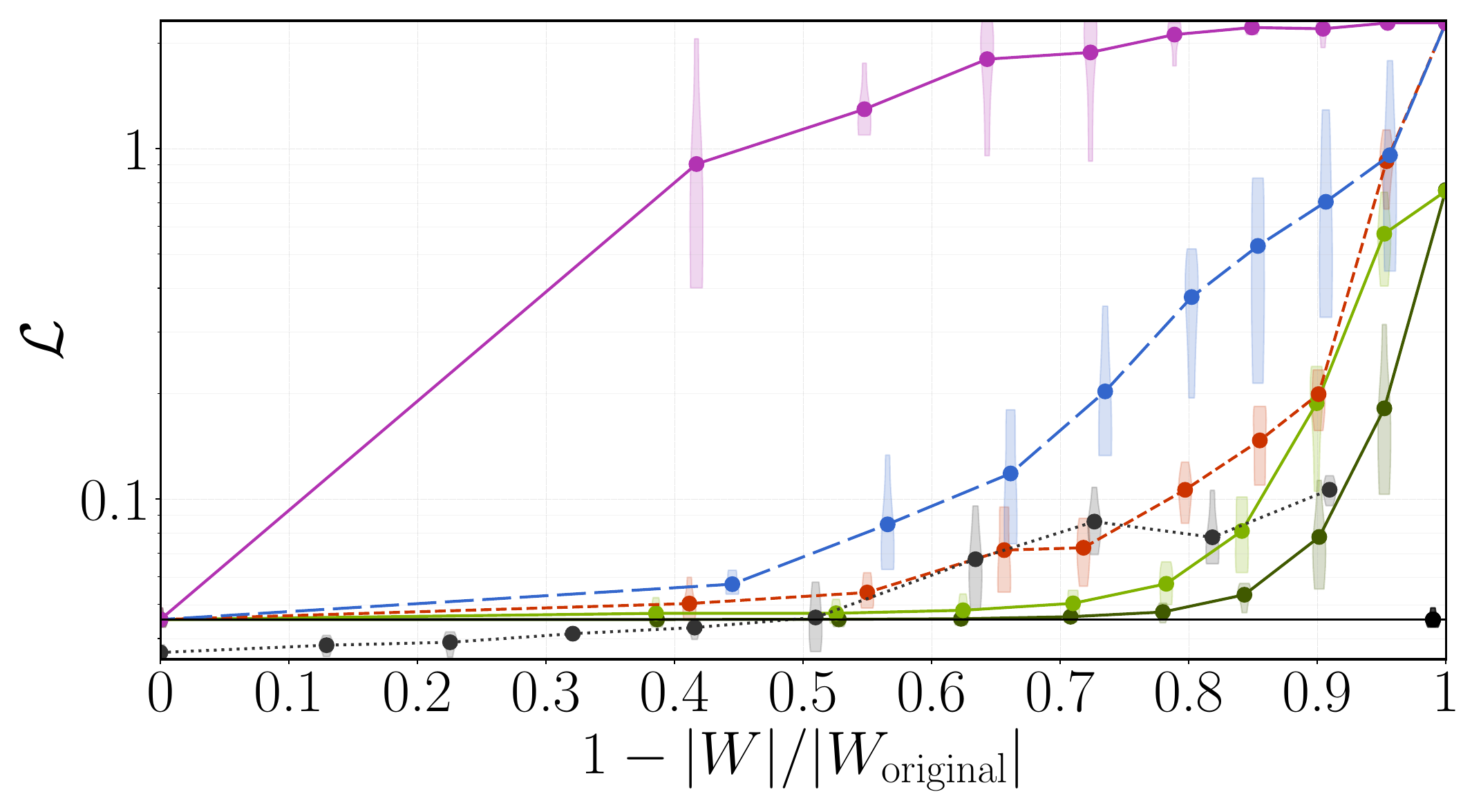}\hspace{5mm}
    \includegraphics[width=0.48\linewidth]{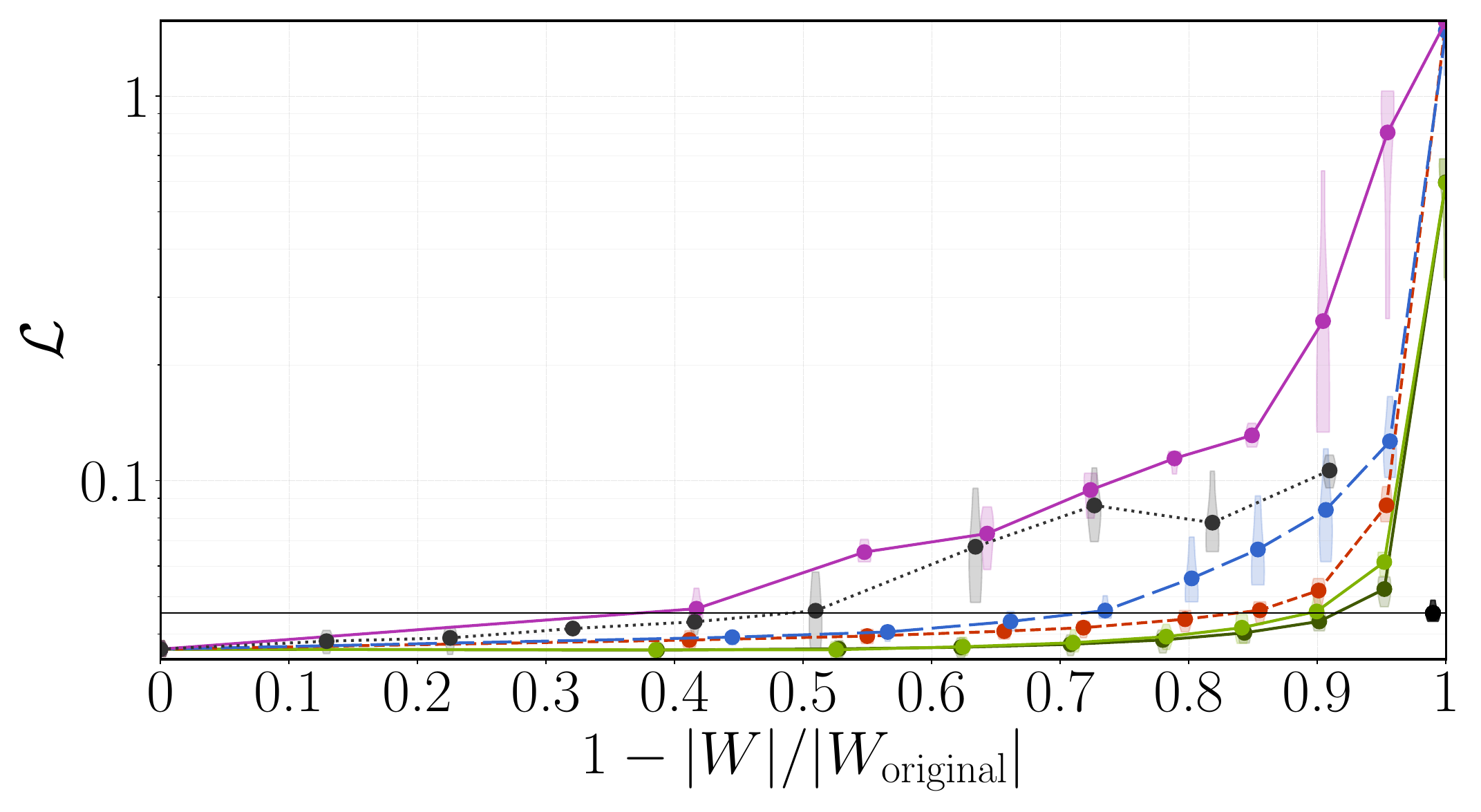} \\
    \centering\textbf{Architecture 2} \\
    \vspace{-2mm}
    \subfloat[Training-pruning]{\includegraphics[width=0.48\linewidth]{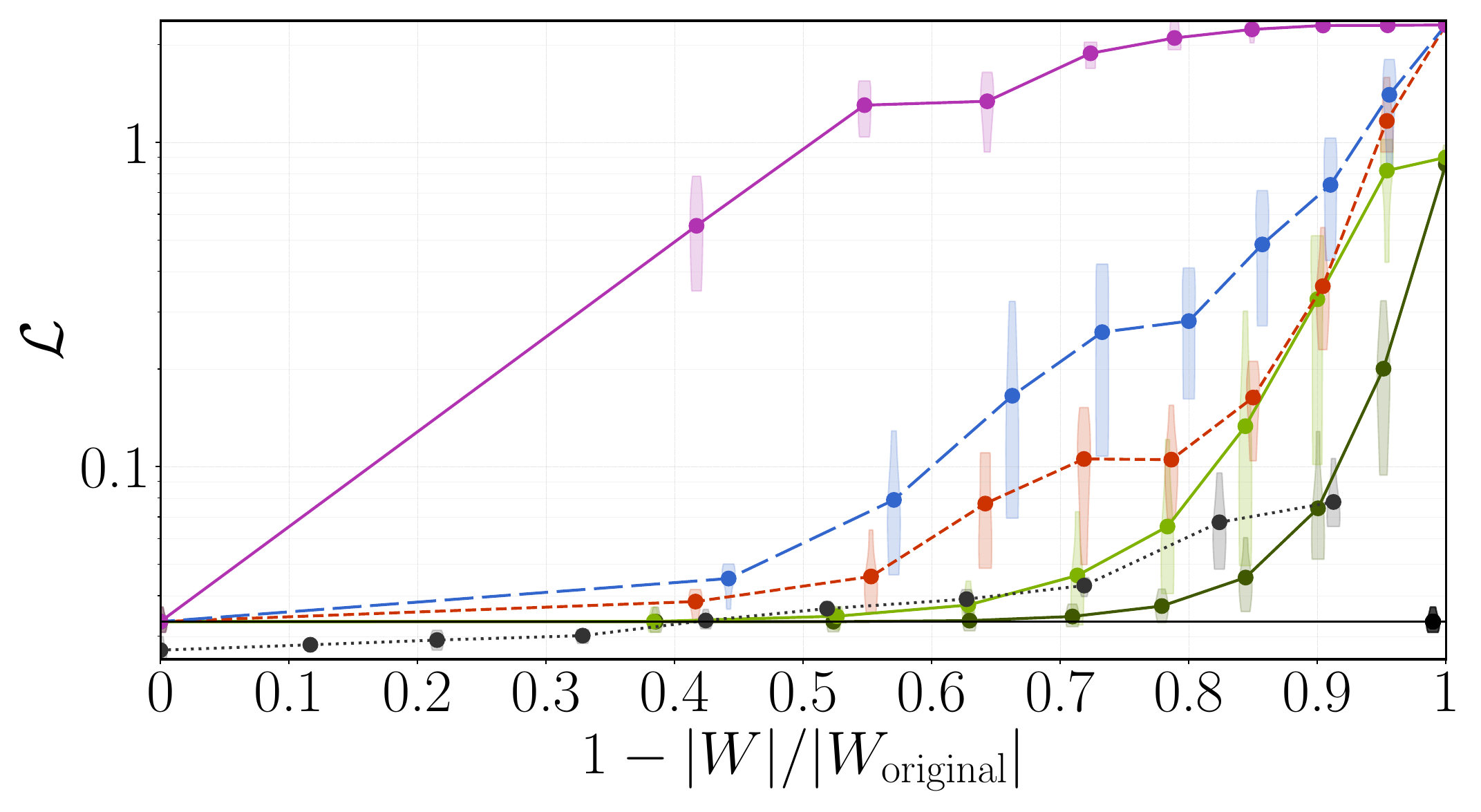}}\hspace{5mm}
    \subfloat[Training–pruning–training]{\includegraphics[width=0.48\linewidth]{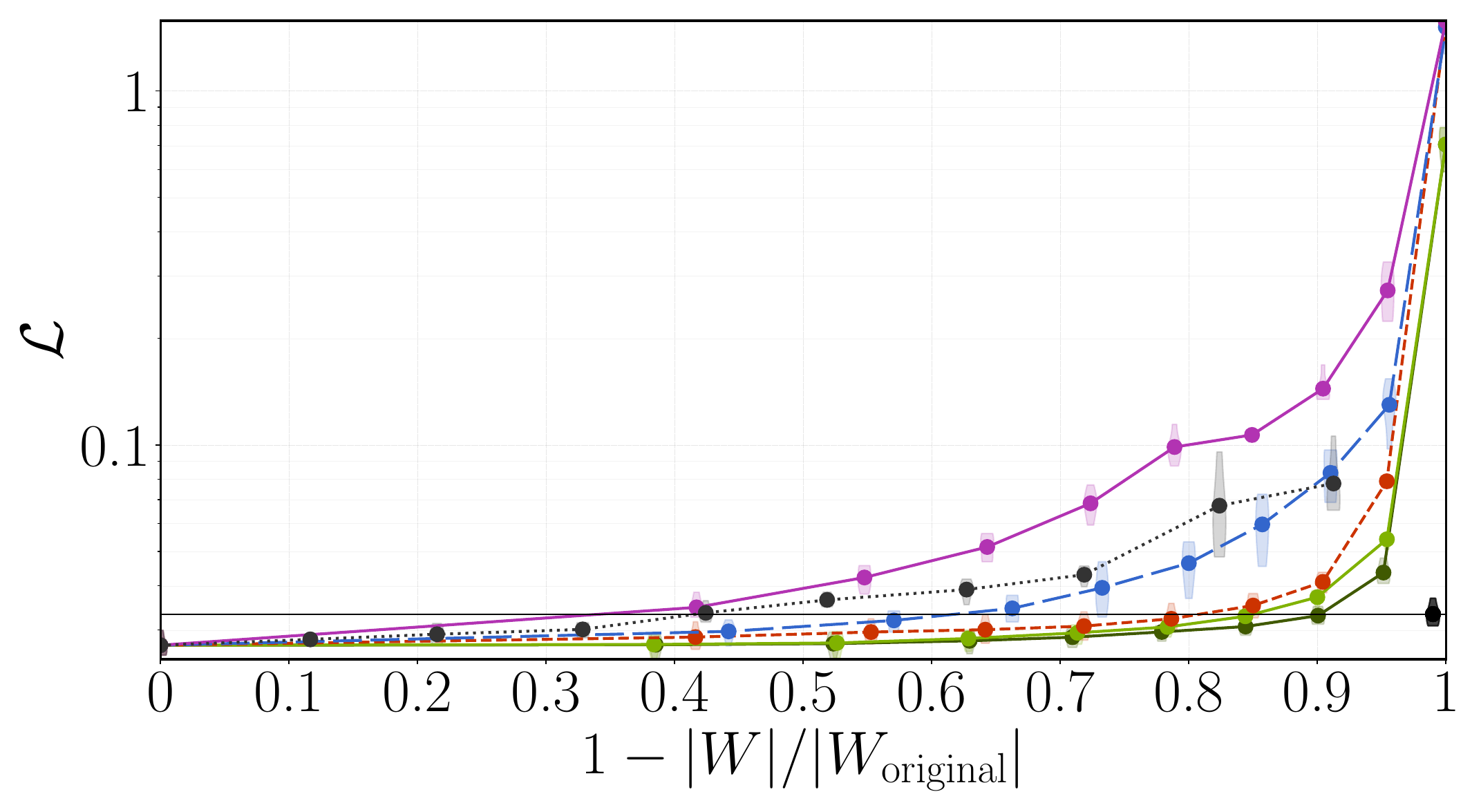}} \\
    \end{minipage} \\ 
    \vspace{5mm}
    \includegraphics[width=0.5\linewidth]{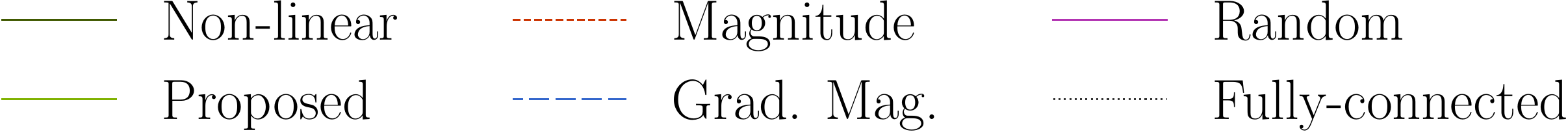}
    \caption{Test losses on MNIST. In the left column, the losses are computed immediately after applying the pruning methods. In the right column, the losses are computed after fine-tuning, so that the overall procedure can be summarized as training–pruning–training.}
    \label{fig:losses_mnist}
\end{figure}

The main results are shown in Fig.~\ref{fig:losses_mnist}, where the loss values are plotted against the pruning ratio for both architectures in Tab.~\ref{tab:architectures_mnist}. All the pruning methods described in the itemized list above are represented, highlighting the average loss across different initializations. Additionally, a violin plot is used for each pruning ratio to illustrate the variability in the results.

For completeness, we report also the baseline loss, i.e., the value computed immediately before pruning, shown as a horizontal line. 

In the left column, the loss values computed immediately after pruning (without any weight adjustment) are shown. Interestingly, using the proposed method, it is possible to remove up to approximately $50\%$ of the weights without a significant change in the loss and without re-training the pruned network.

In the right column, the loss values computed after fine-tuning are shown. We observe that the proposed method consistently achieves lower loss than all other methods, with the loss remaining nearly unchanged even with $80$--$90\%$ of the weights removed.

\begin{table}[h!]
	\centering
	\begin{tabular}{lll}
		& \textbf{layer's size}        & \textbf{$|W|$}  \\
		\hline \\
		\textbf{Architecture 1}& 784, 32, 32, 10     & \numprint{26432} \\
		\textbf{Architecture 2}& 784, 64, 64, 10     & \numprint{54912} \\                               
	\end{tabular}
	\caption{Baseline neural network architectures for MNIST. $|W|$ is the total number of trainable weights.}
	\label{tab:architectures_mnist}
\end{table}

\clearpage

\subsection{Case 2: Partial differential equation with noisy data}
\begin{figure*}
	\centering
	\includegraphics[width=0.3\linewidth]{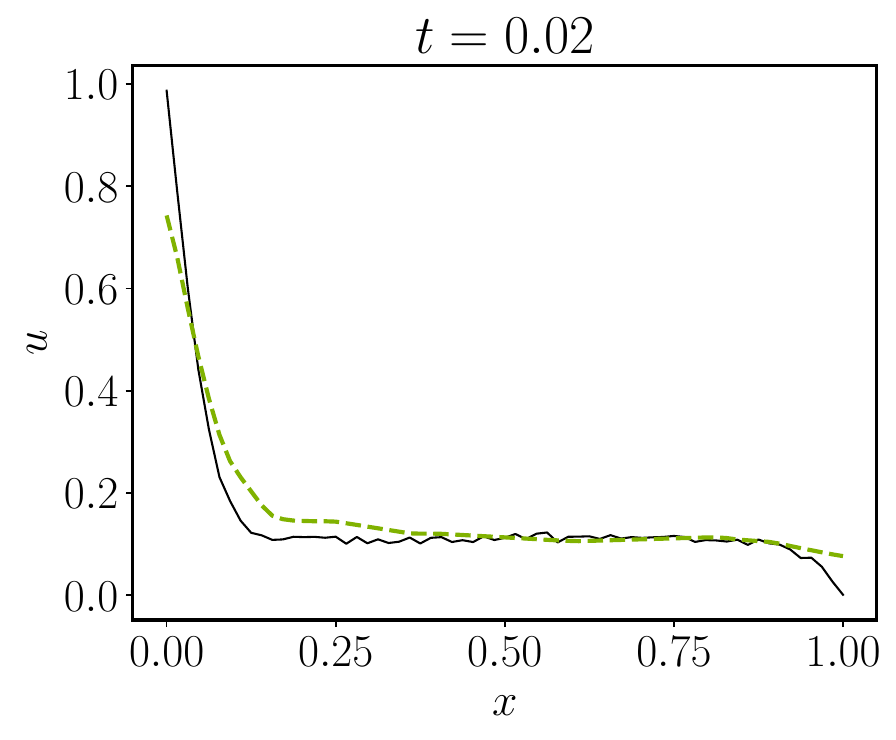}\hspace{2mm}
	\includegraphics[width=0.3\linewidth]{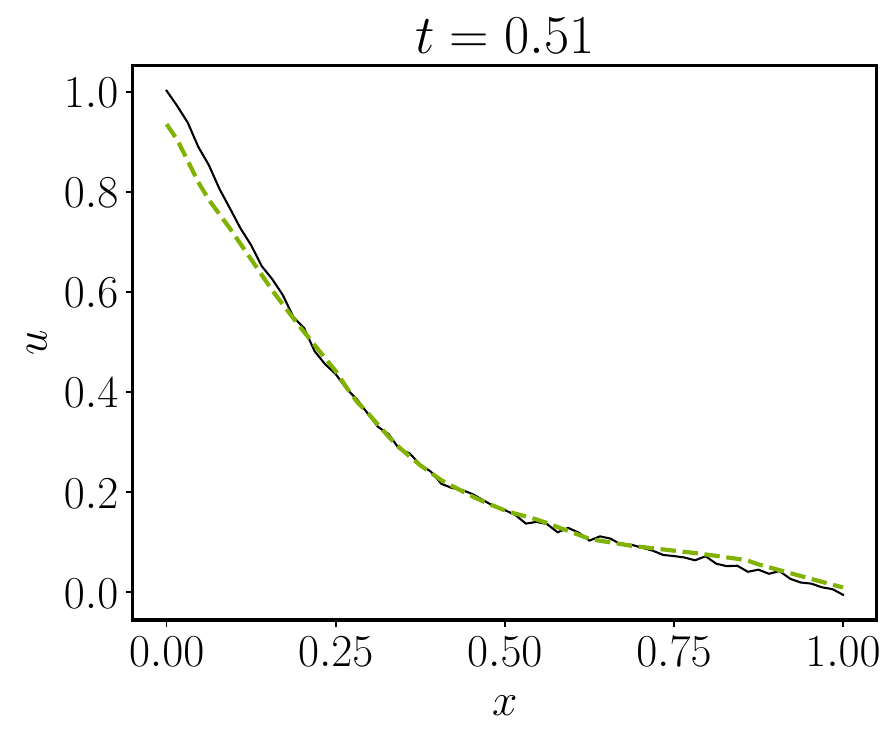}\hspace{2mm}
	\includegraphics[width=0.3\linewidth]{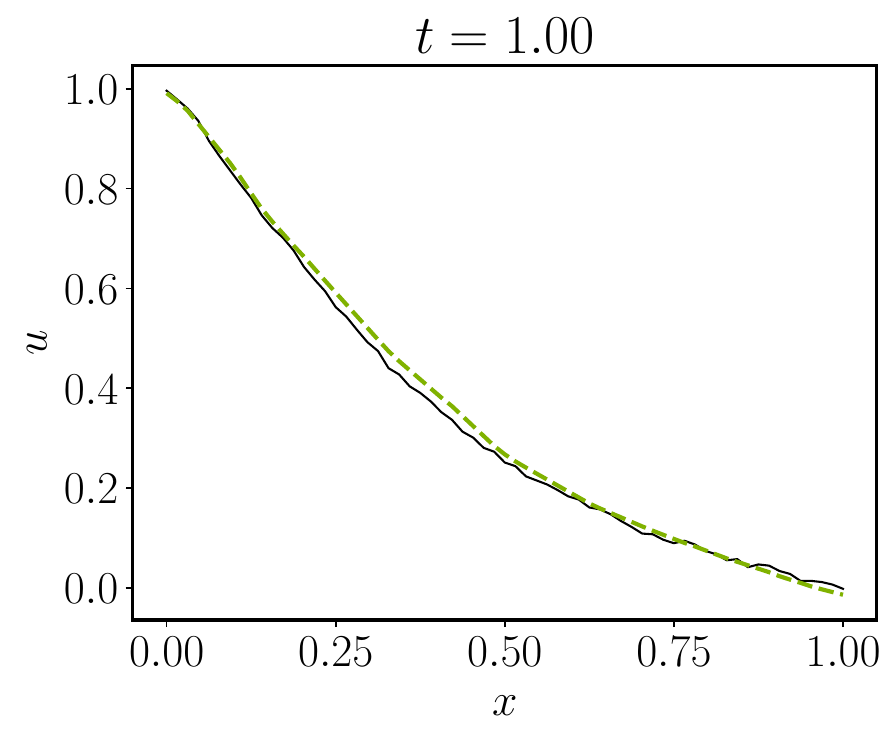}
	\caption{Time snapshots of $u(x,t)$. The noisy data with $d \sim \mathcal{U}(-0.005, 0.005)$ are shown in black, while the green dashed line represents $u(x,t)$ computed by the pruned neural network with architecture~2 (see Tab.~\ref{tab:architectures_pde}). In all panels, the network is pruned using the proposed method with a pruning ratio of $0.7$.}
	\label{fig:pde}
\end{figure*}

In this case, the dataset is derived from the benchmark described in \cite{Takamoto2022}. The data represent the solution $u$, computed on a uniform grid, of a 1D partial differential equation. The equations describe a time-dependent diffusion-sorption nonlinear system. The equations are parametric, with the parameters denoted by $\mu$. See \cite{Takamoto2022} for further details. 

The input to the neural networks consists of the vector $(\mu, t, x, u_0)$, where $t$ is time, $x$ is the spatial coordinate, and $u_0$ denotes the initial conditions on the grid. The output of the networks is $u(t,x)$ for the given $\mu$ and $u_0$. 

The dataset used in this paper is smaller than the original one presented in \cite{Takamoto2022} due to the large amount of tests required to compare the methods. Specifically, we restrict the discretization to $10$ of the $100$ time steps, $64$ of the $1024$ grid points, and the first $100$ samples in the parameter space of size $\numprint{10000}$.

To the original dataset, a uniformly distributed noise, $d(t,x)$, is added to make the test more challenging and to assess the robustness of the pruning method. Considering that the solution $u$ spans a range of approximately $[0,1]$, we test the methods with three levels of noise: $d = 0$, $d \sim \mathcal{U}(-0.005, 0.005)$, and $d \sim \mathcal{U}(-0.01, 0.01)$. See also Fig.~\ref{fig:pde} for a graphical visualization. A practical interpretation of the noise can, in principle, be related to a measurement noise or numerical inaccuracies. 

The pruning method is applied to two neural network architectures: one with $3$ hidden layers, and the other with $6$ hidden layers, as summarized in Tab.~\ref{tab:architectures_pde}.

\begin{figure}[H]
    \centering
    \begin{minipage}{1\textwidth}
    \centering{\textbf{Architecture 1}} \\
    \begin{turn}{90}
        \hspace{1cm} $Noise = 0$
    \end{turn}
    \vspace{3mm}
    \includegraphics[width=0.48\linewidth]{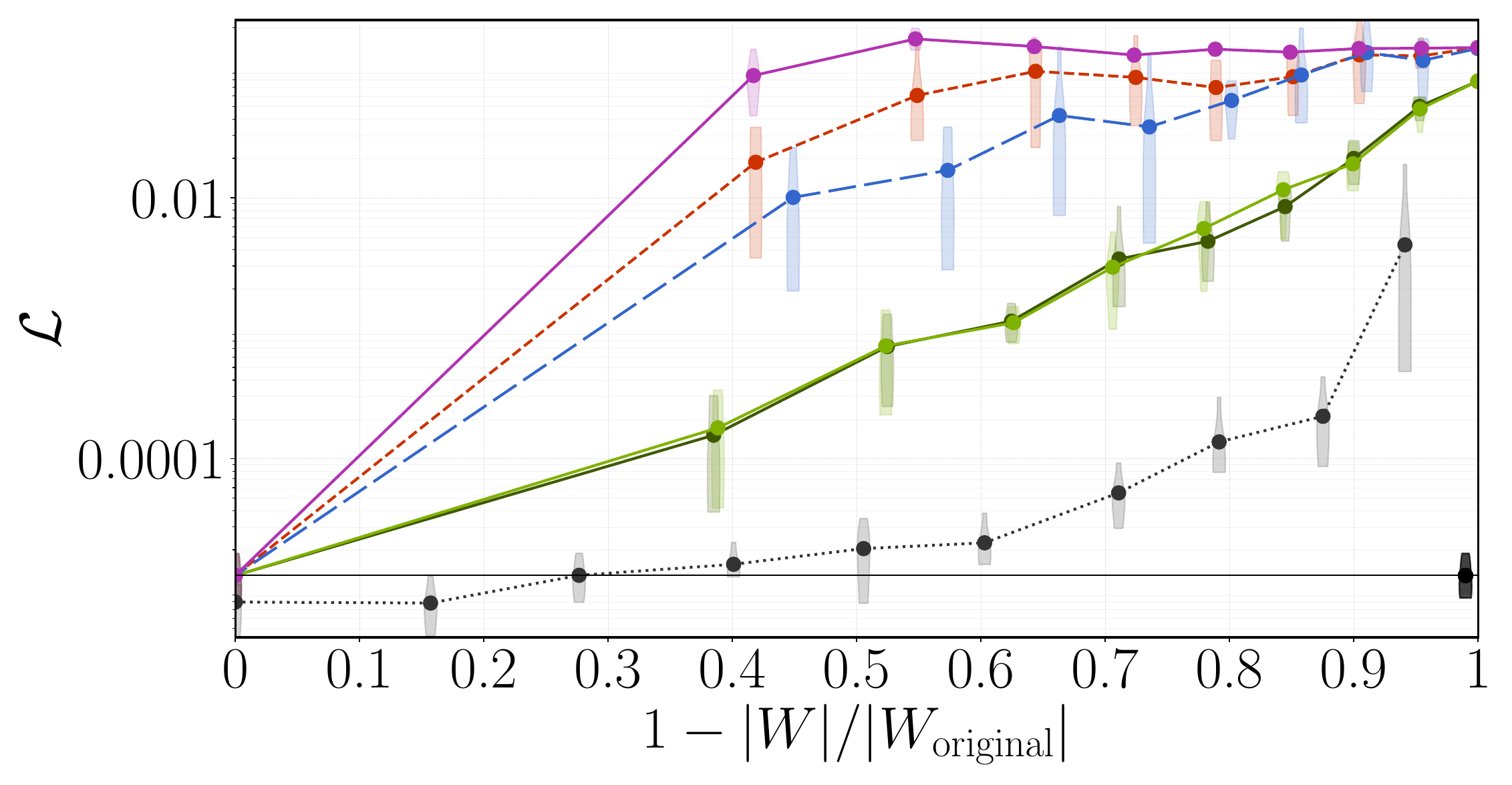}
    \includegraphics[width=0.48\linewidth]{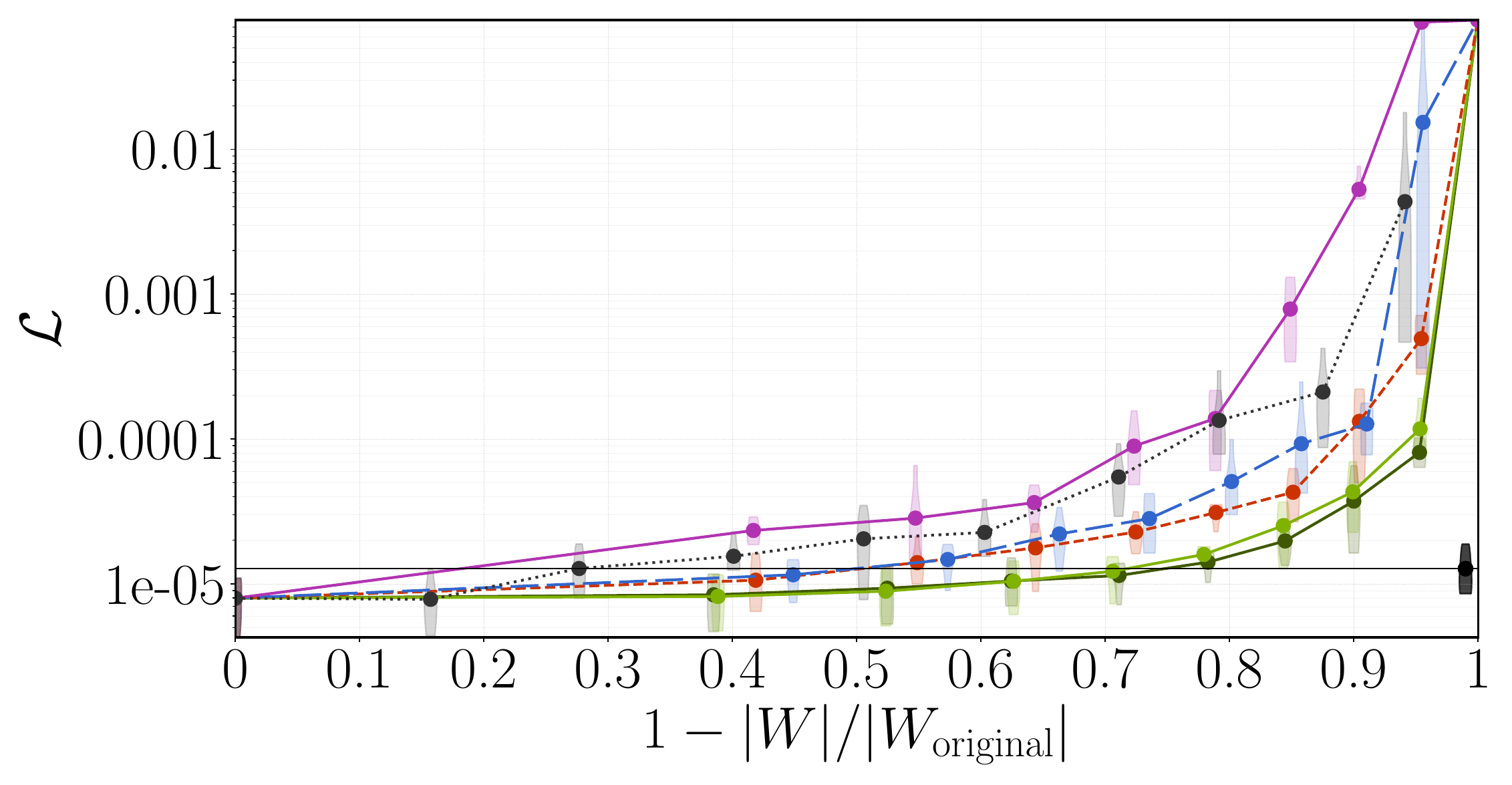} \\
    \begin{turn}{90}
        \hspace{1cm} $Noise = 0.005$
    \end{turn}
    \includegraphics[width=0.48\linewidth]{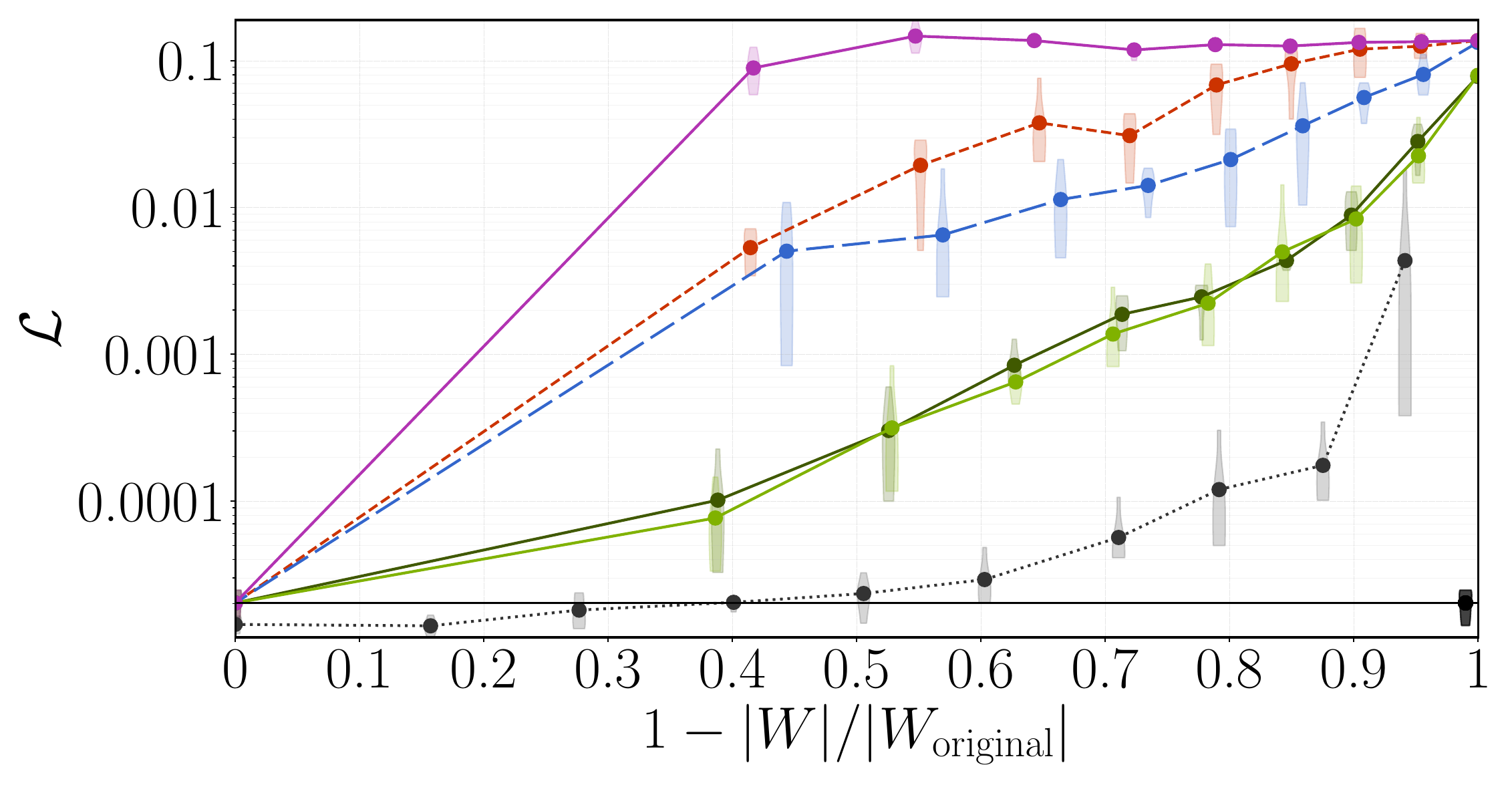}
    \includegraphics[width=0.48\linewidth]{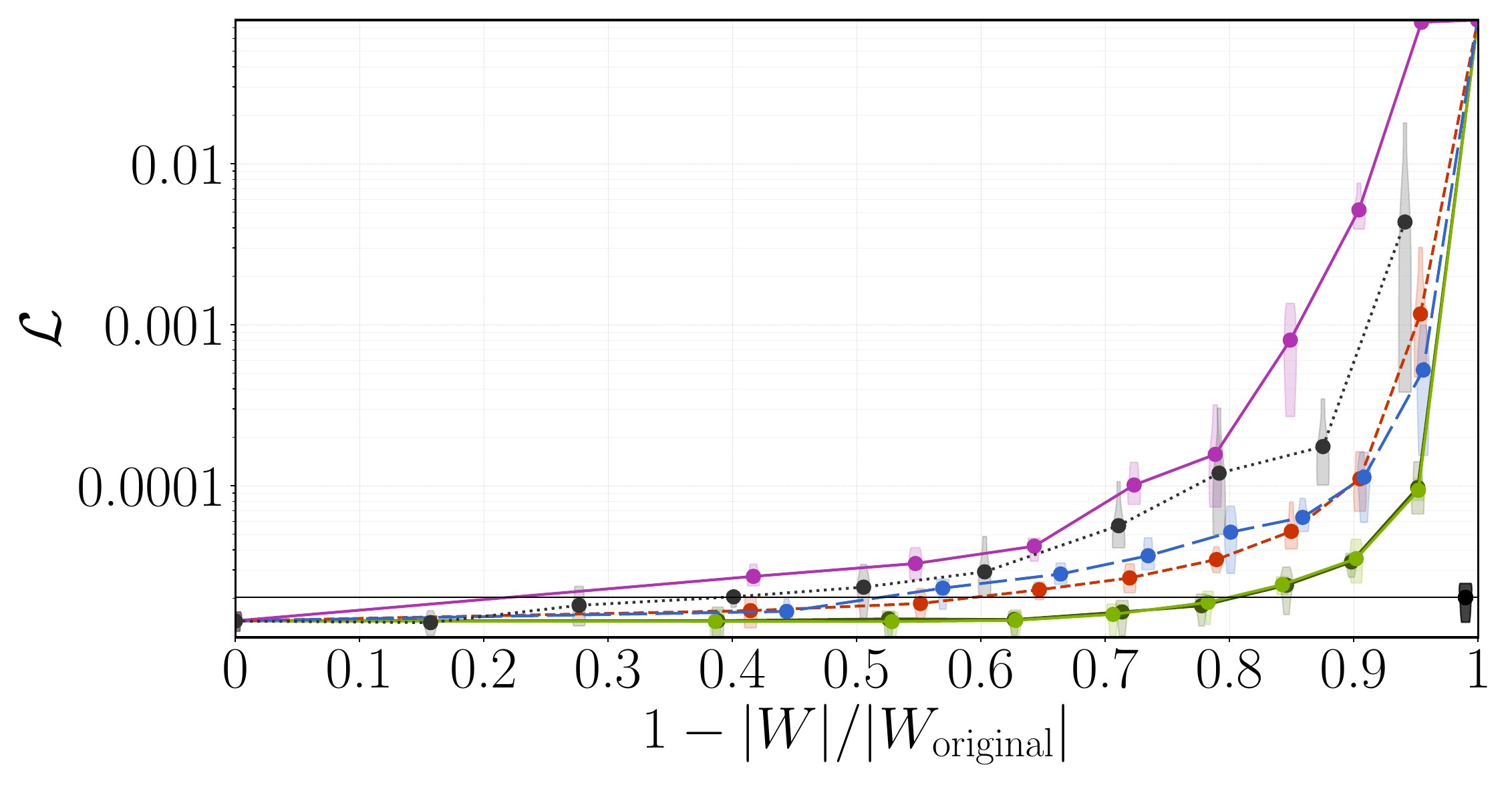} \\
    \begin{turn}{90}
        \hspace{1cm} $Noise = 0.01$
    \end{turn}
    \subfloat[Training–pruning]{\includegraphics[width=0.48\linewidth]{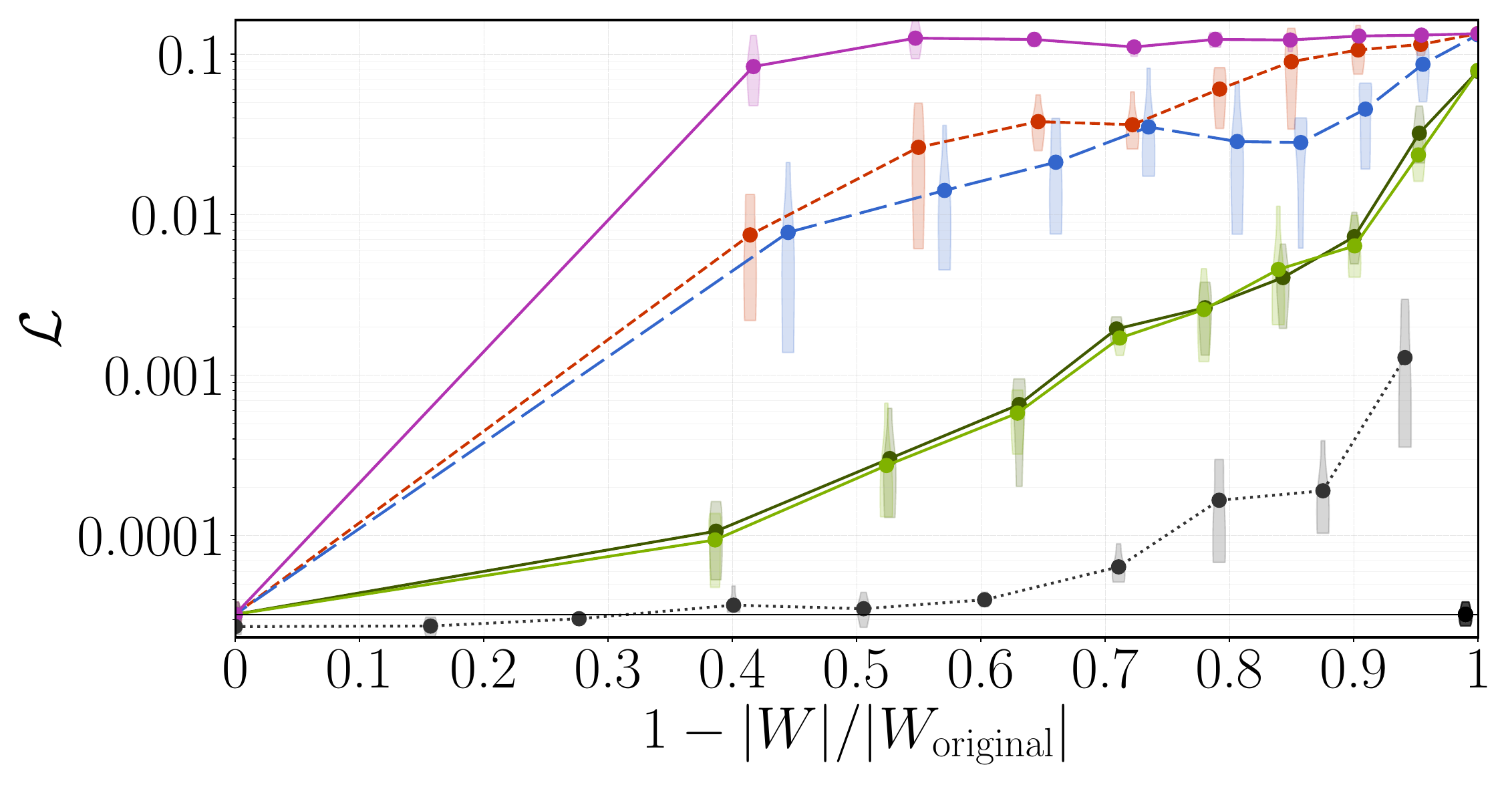}}
    \subfloat[Training–pruning–training]{\includegraphics[width=0.48\linewidth]{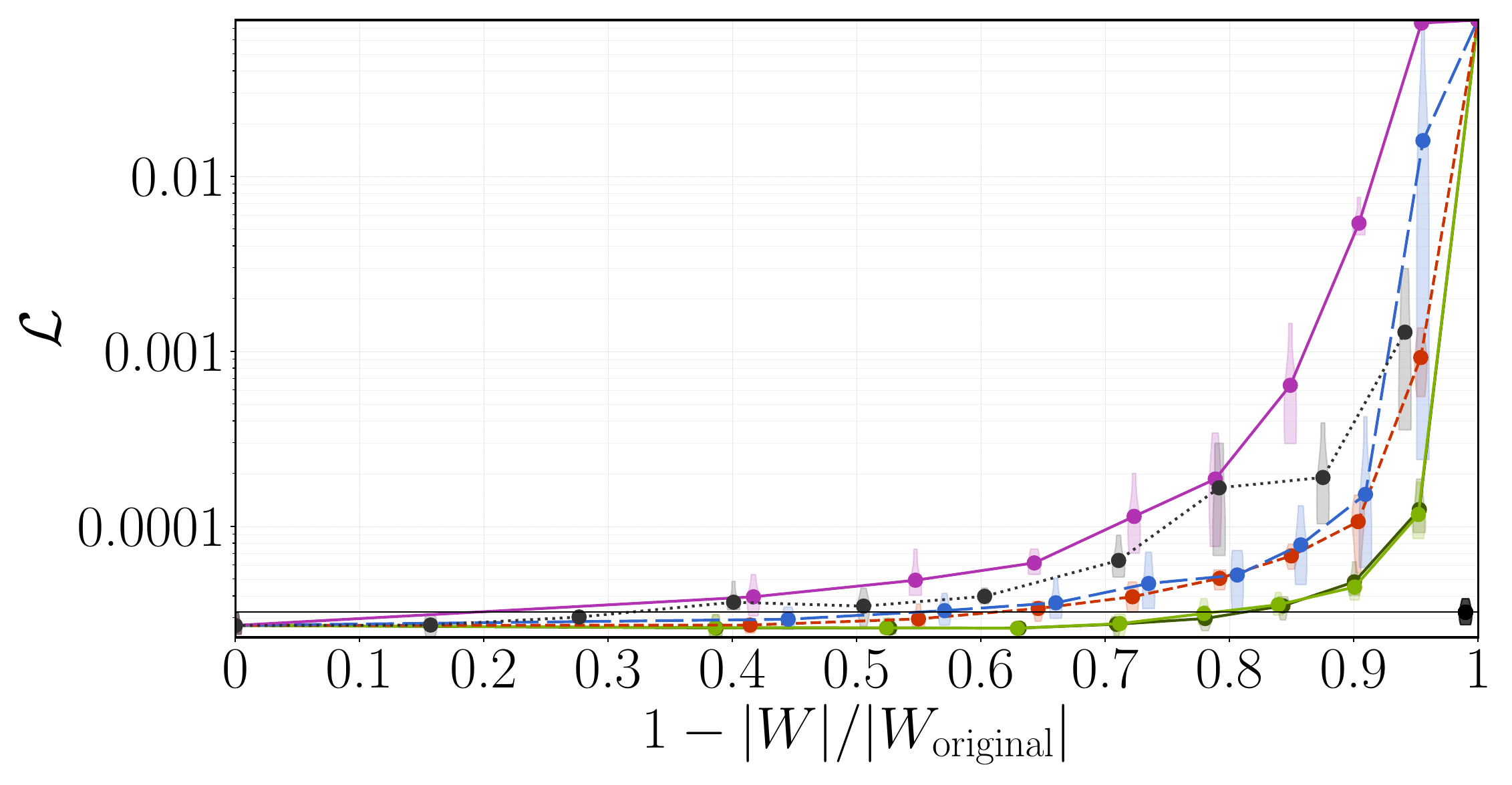}} \\
    \end{minipage} \\
    \vspace{5mm}
    \includegraphics[width=0.5\linewidth]{fig/losses_legend.pdf}
    \caption{Test losses on Diffusion-Sorption PDE. In the left column, the losses are computed immediately after applying the pruning methods. In the right column, the losses are computed after fine-tuning, so that the overall procedure can be summarized as training–pruning–training. }
    \label{fig:losses_pde_1}
\end{figure}

\begin{figure}[H]
    \centering
    \begin{minipage}{1\textwidth}
    \centering
    \textbf{Architecture 2} \\
    \begin{turn}{90}
        \hspace{1cm} $Noise = 0$
    \end{turn}
    \vspace{3mm}
    \includegraphics[width=0.48\linewidth]{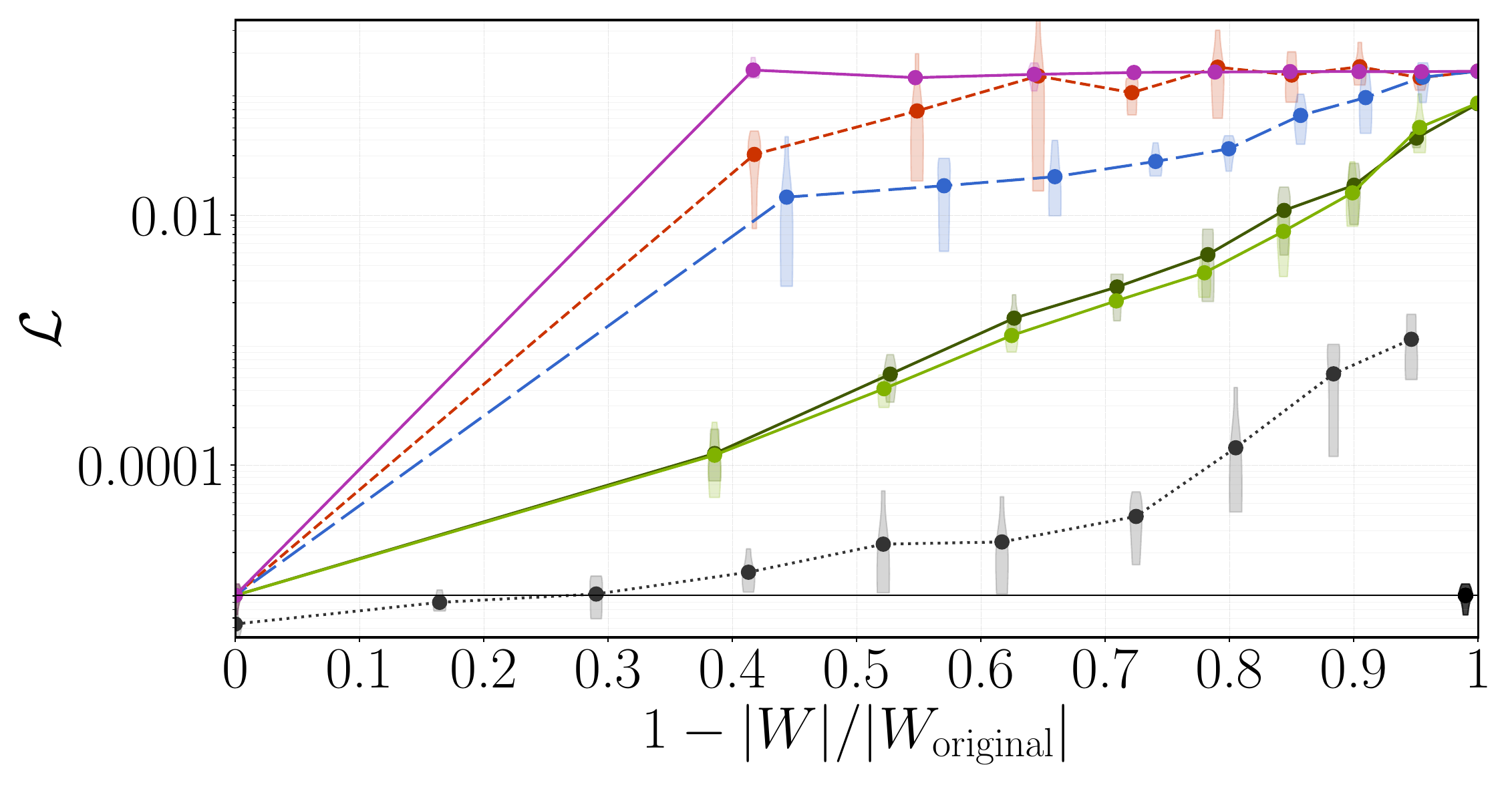}
    \includegraphics[width=0.48\linewidth]{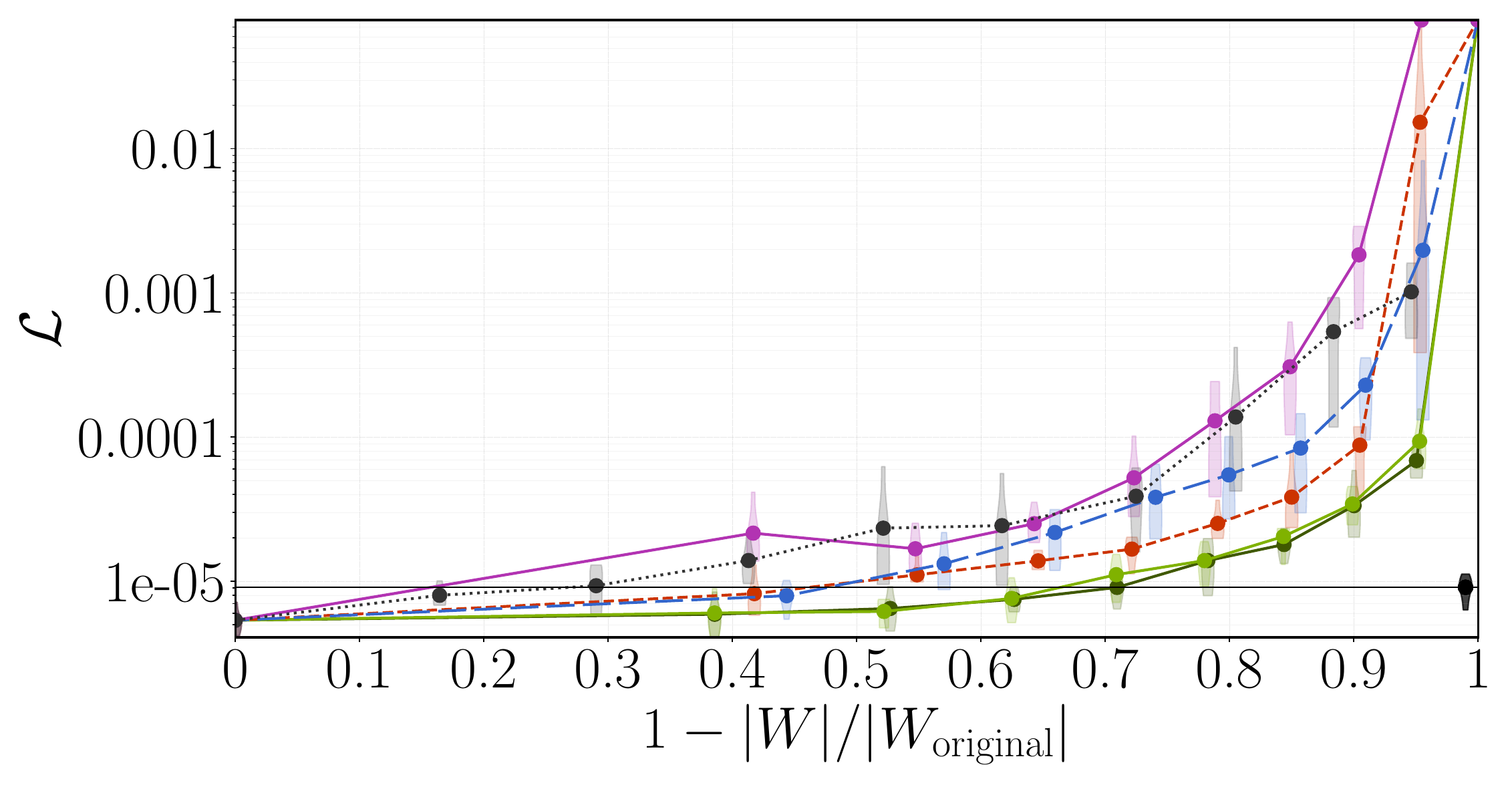} \\
    \begin{turn}{90}
        \hspace{1cm} $Noise = 0.005$
    \end{turn}
    \includegraphics[width=0.48\linewidth]{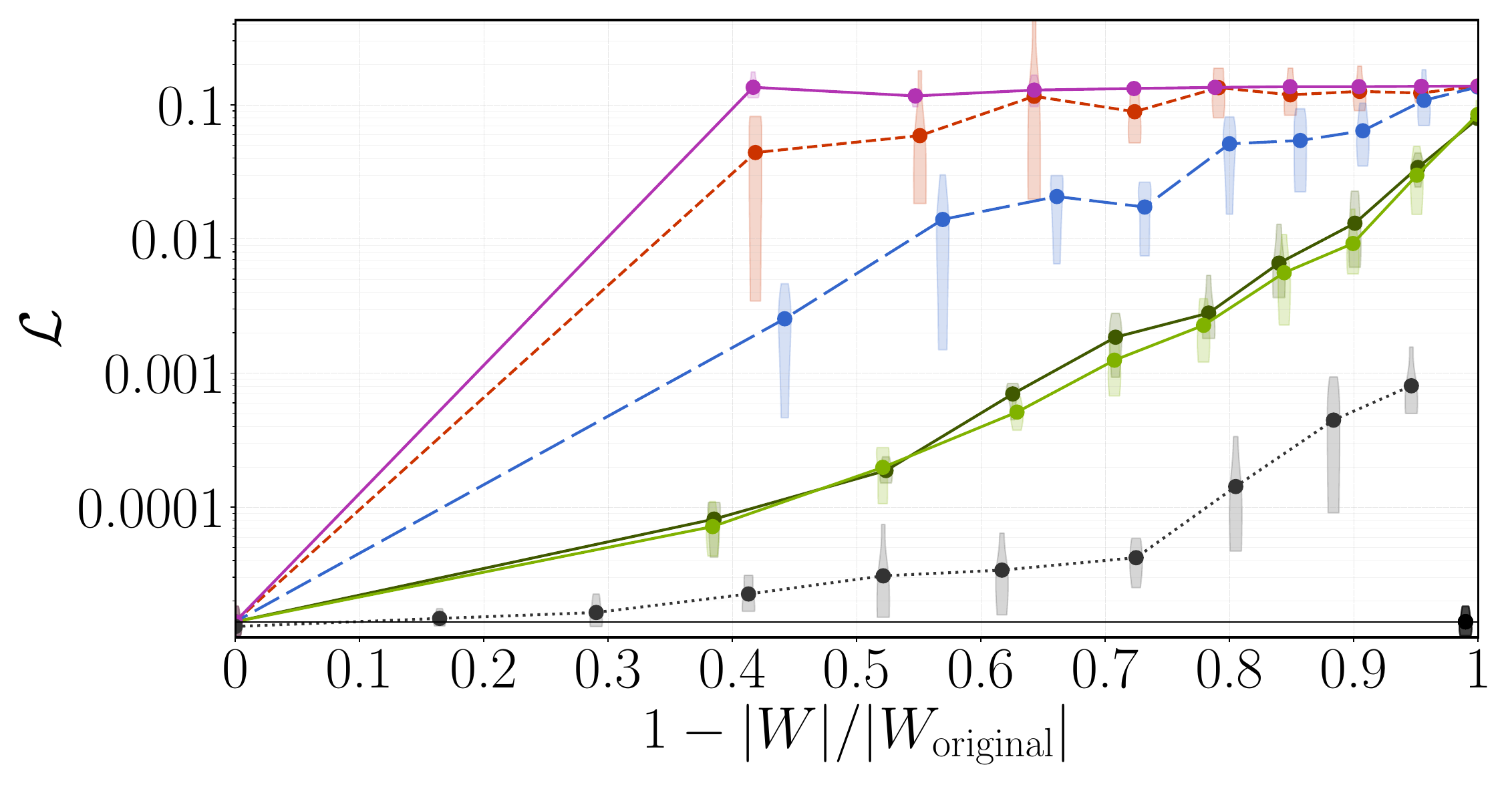}
    \includegraphics[width=0.48\linewidth]{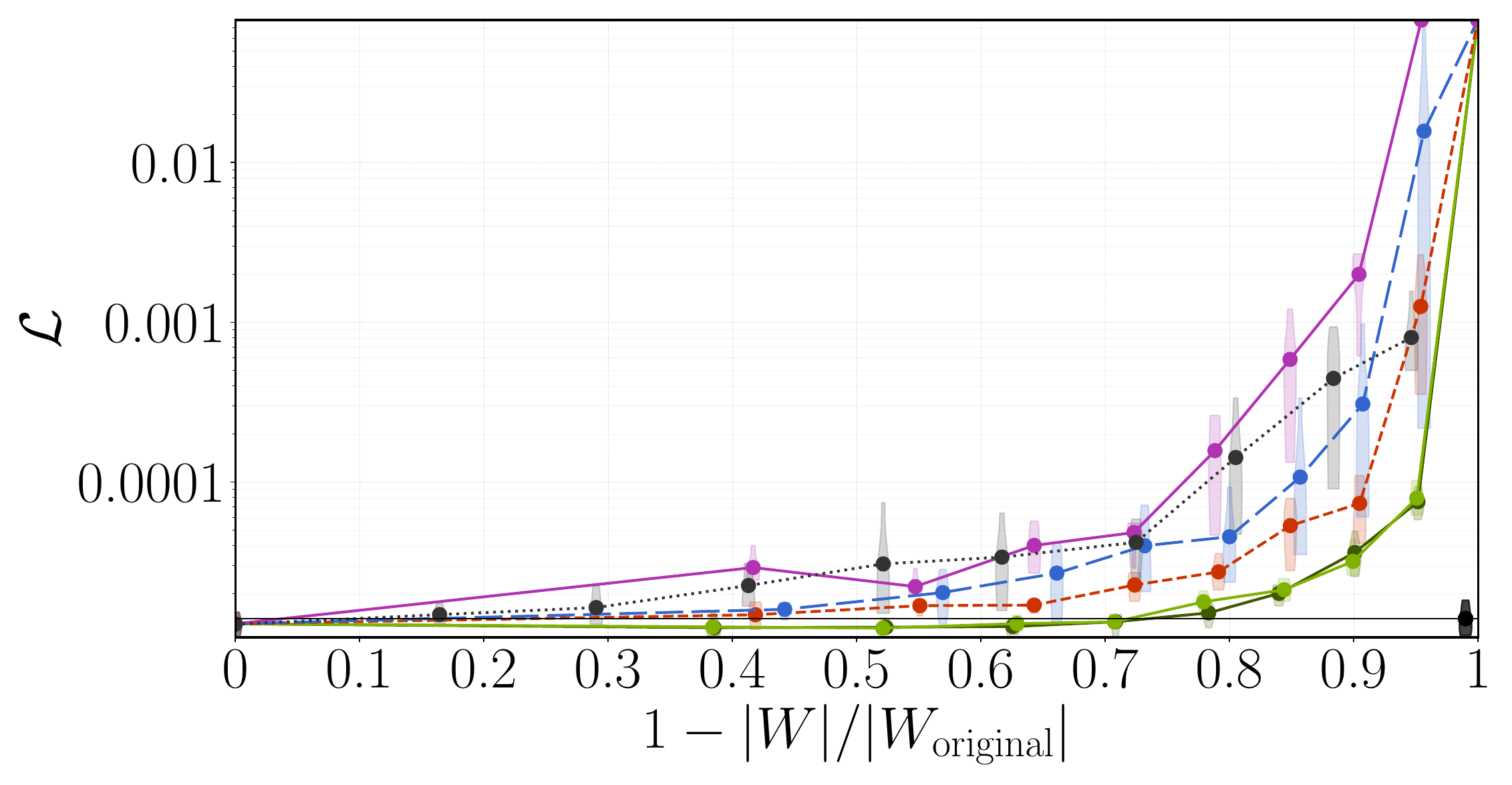} \\
    \begin{turn}{90}
        \hspace{1cm} $Noise = 0.01$
    \end{turn}
    \subfloat[Training–pruning]{\includegraphics[width=0.48\linewidth]{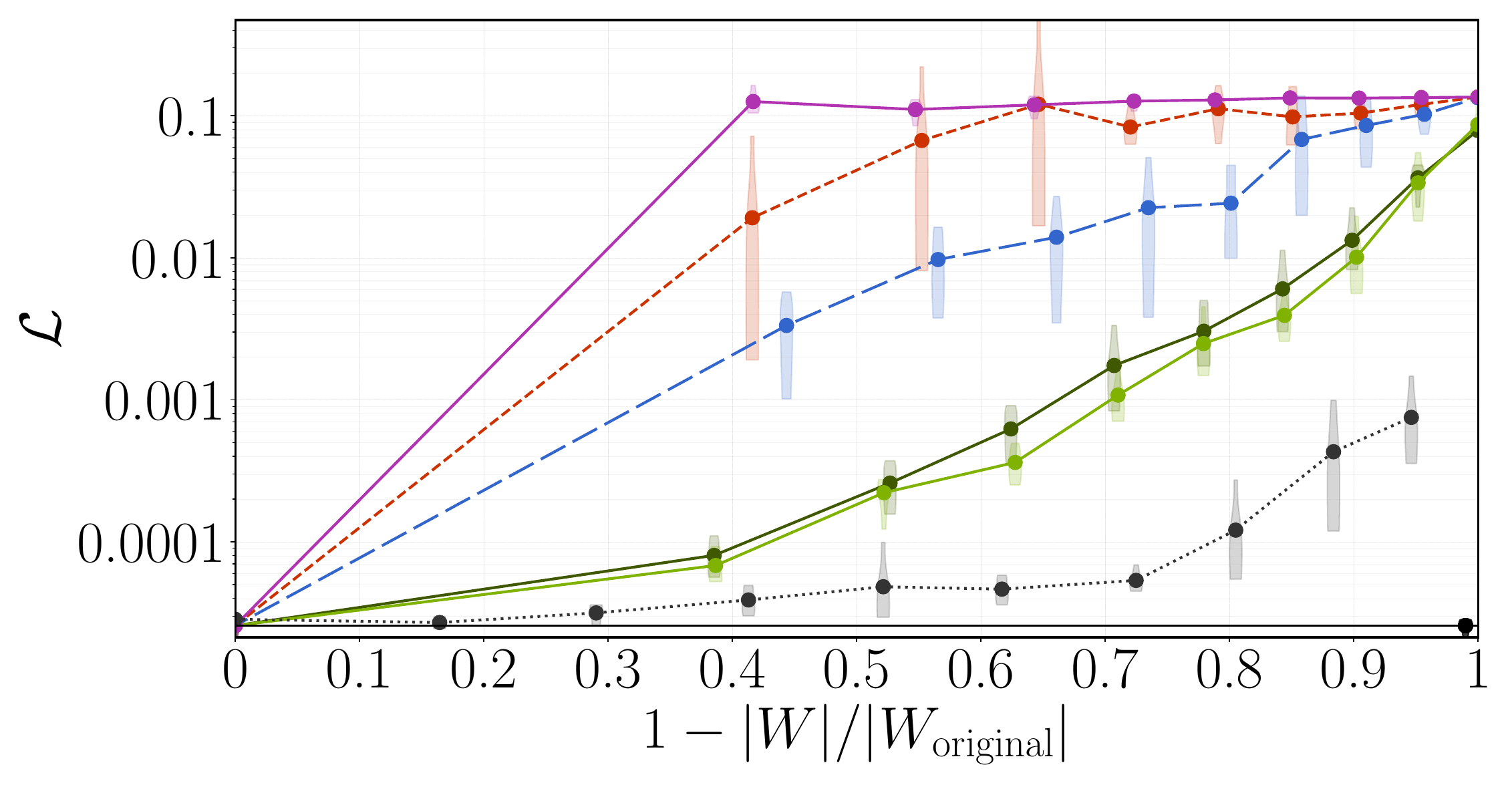}}
    \subfloat[Training–pruning–training]{\includegraphics[width=0.48\linewidth]{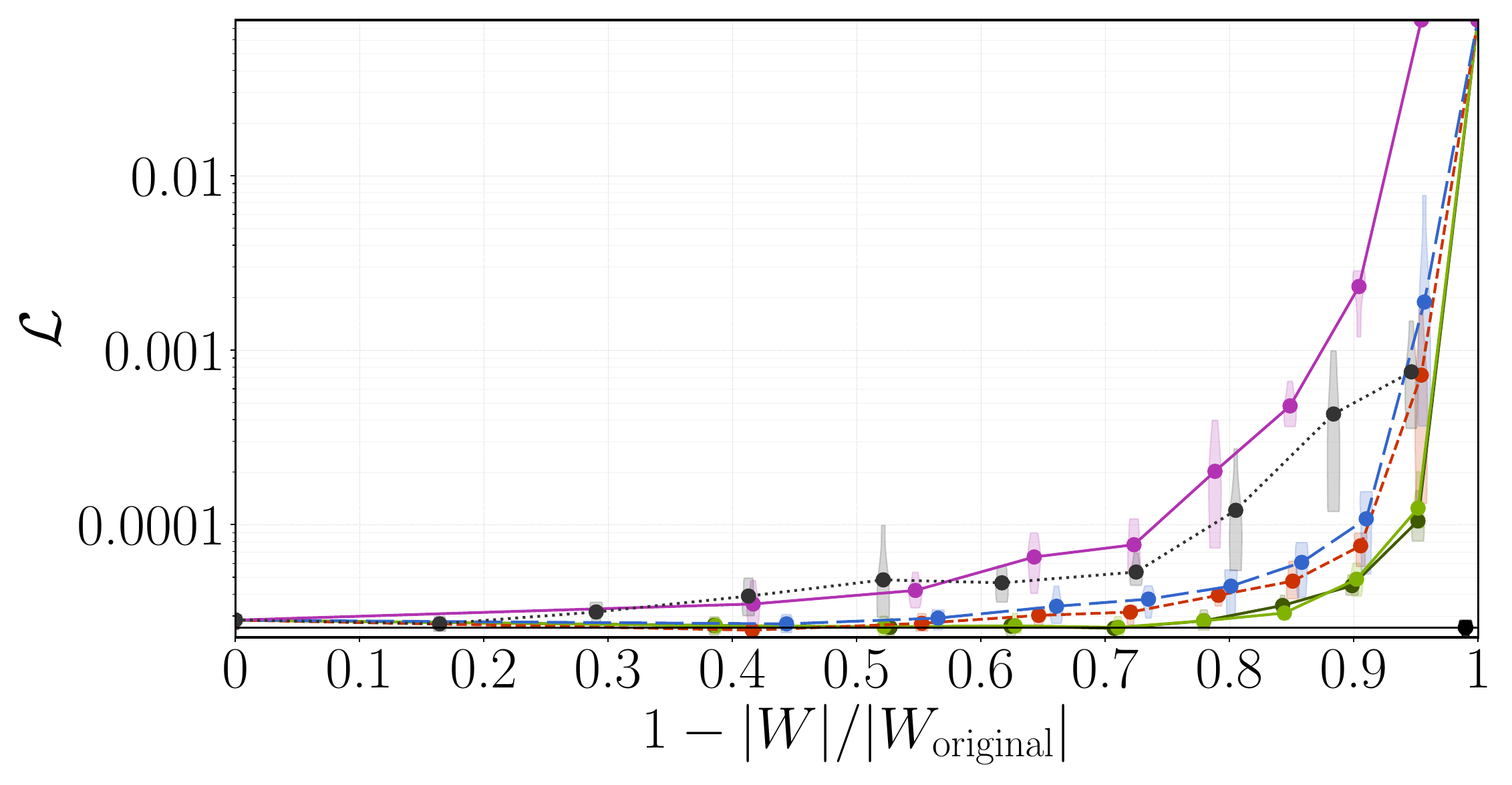}} \\
    \end{minipage}\\
    \vspace{5mm}
    \includegraphics[width=0.5\linewidth]{fig/losses_legend.pdf}
    \caption{Test losses on Diffusion-Sorption PDE. The bottom row corresponds to a high noisy data. In this case, it is practically difficult to reach low loss values during training, so that the first training phase is sufficient to reach the practical minimum, and the second training phase does not significantly improve performance. Therefore, it is possible for the loss of the baseline to be comparable to that of the fully trained networks, as observed in the bottom-right panel.}
    \label{fig:losses_pde_2}
\end{figure}

\begin{table}[H]
	\centering
	\begin{tabular}{lll}
		& \textbf{layer's size}        & \textbf{$|W|$}  \\
		\hline \\
		\textbf{Architecture 1}& 68, 32, 32, 32, 1 & \numprint{4256}   \\
		\textbf{Architecture 2}& 68,  64, 32, 32, 16, 16, 1 & \numprint{8208}   \\                                  
	\end{tabular}
	\caption{Baseline neural network architectures for Diffusion-Sorption. $|W|$ is the total number of trainable weights.}
	\label{tab:architectures_pde}
\end{table}
Figs.~\ref{fig:losses_pde_1}, \ref{fig:losses_pde_2} show the test loss values versus the pruning ratio for both architectures listed in Tab.~\ref{tab:architectures_pde} and for the three aforementioned levels of noise. We observe that, overall, all pruning strategies are robust with respect to noise. The proposed method consistently outperforms all other methods, except for the non-linear strategy, which is occasionally slightly better. However, it should be noted that the non-linear method is roughly $|W|$ times more computationally expensive than the proposed method. The latter has a computational cost comparable to that of a single backpropagation step and therefore requires a negligible amount of time when compared to the overall training time. Here, we do not provide timing comparisons, as they are heavily influenced by the hardware and code optimizations, which are beyond the scope of this work.


\clearpage

\section{Conclusions}\label{sec:conclusion}

This work focused on the presentation of a novel pruning strategy, that we implemented considering an ad-hoc weight importance quantification function, based not only on the estimated effect of weight removal but also on the collateral contribution that an optimally tuned adjacent bias could give. Experimental results show a significant improvement against classical pruning strategies, suggesting that the interlaced action of parameters may be important in constructing effective compression techniques. We claim that this should not be too surprising: parameters of any neural network exist as a realization of a highly correlated random vector, that is the image of the initialization through the optimization procedure. This simple observation suggests that correlation between parameters (in our case, weights and biases) should not be neglected, and their consideration (for instance, with our compensation mechanism) could lead to high quality parameter compression techniques. In particular, while ``atomic'' weight importance metrics such as weight magnitude constitute a very efficient and sometimes effective strategy, we argue that the complex representation of information in modern neural networks requires non myopic approaches, possibly considering the effect of long range correlation between weights in different layers. This second idea is at the basis of the construction of the $\Delta$ function: a Taylor based expansion recovers the mechanics of the prediction for asymptotically small displacements of the weights, considering the effect across all the successive subnetwork. As a final point, we also would like to position our idea of using the output function rather than the loss function for constructing our importance metrics, despite the latter being certainly more frequent in literature. In particular, we observe that, at perfect convergence, the derivative of the loss function should be identically $0$. Consequently, the ranking induced by loss-gradient-based metrics is dominated by the effects of numerical noise. We conjecture that this partially explains the behavior of the  pruning technique based on the loss gradient and weight magnitude ($\mathcal I_{W^{\ell}_{ij}} = \partial_{W^{\ell}_{ij}} \mathcal L$) presented in the experiments, which has poorer performance than our proposed method. This is likely due to the influence of residuals, which vanish at convergence, thereby eliminating the advantage of such an approach.

In this work, we focus exclusively on real-valued fully connected neural networks, which constitute a fundamental general-purpose architecture. Although not reported here, based on encouraging results obtained, the proposed method may be extended to other architectures after appropriate studies.

\section*{Acknowledgements}
The present research has received support from the project FIS, MUR, Italy 2025-2028, Project code: FIS-2023-02228, CUP: D53C24005440001, "SYNERGIZE: Synergizing Numerical Methods and Machine Learning for a new generation of computational models".
L.M. and F.R. have received support from the project PRIN2022, MUR, Italy, 2023-2025, P2022N5ZNP “SIDDMs: shape-informed data-driven models for parametrized PDEs, with application to computational cardiology”, funded by the European Union (Next Generation EU, Mission 4 Component 2).
The authors of this work acknowledge the grant Dipartimento di Eccellenza 2023-2027, MUR, Italy.
A.F., A.S., and F.R. are members of GNCS, ``Gruppo Nazionale per il Calcolo Scientifico'' (National Group for Scientific Computing) of INdAM (Istituto Nazionale di Alta Matematica).

\clearpage
\bibliographystyle{abbrv}
\bibliography{all}

@InProceedings{LeCun1989,
  author    = {LeCun, Yann and Denker, John and Solla, Sara},
  booktitle = {Advances in Neural Information Processing Systems},
  title     = {Optimal Brain Damage},
  year      = {1989},
  volume    = {2},
  comment   = {https://proceedings.neurips.cc/paper/1989/file/6c9882bbac1c7093bd25041881277658-Paper.pdf},
}

@Article{Kingma2014,
  author    = {Kingma, Diederik P. and Ba, Jimmy},
  journal   = {arXiv:1412.6980},
  title     = {Adam: A Method for Stochastic Optimization},
  year      = {2014},
  copyright = {arXiv.org perpetual, non-exclusive license},
  doi       = {10.48550/ARXIV.1412.6980},
  publisher = {arXiv},
}

@Article{Pant2021,
  author    = {Pant, Pranshu and Doshi, Ruchit and Bahl, Pranav and Barati Farimani, Amir},
  journal   = {Physics of Fluids},
  title     = {Deep learning for reduced order modelling and efficient temporal evolution of fluid simulations},
  year      = {2021},
  issn      = {1089-7666},
  number    = {10},
  volume    = {33},
  doi       = {10.1063/5.0062546},
  publisher = {AIP Publishing},
}

@Article{Lu2021,
  author    = {Lu, Lu and Jin, Pengzhan and Pang, Guofei and Zhang, Zhongqiang and Karniadakis, George Em},
  journal   = {Nature Machine Intelligence},
  title     = {Learning nonlinear operators via {DeepONet} based on the universal approximation theorem of operators},
  year      = {2021},
  issn      = {2522-5839},
  month     = mar,
  number    = {3},
  pages     = {218--229},
  volume    = {3},
  doi       = {10.1038/s42256-021-00302-5},
  publisher = {Springer Science and Business Media LLC},
}

@Article{Raissi2019,
  author    = {Raissi, M. and Perdikaris, P. and Karniadakis, G.E.},
  journal   = {Journal of Computational Physics},
  title     = {Physics-informed neural networks: A deep learning framework for solving forward and inverse problems involving nonlinear partial differential equations},
  year      = {2019},
  issn      = {0021-9991},
  pages     = {686--707},
  volume    = {378},
  doi       = {10.1016/j.jcp.2018.10.045},
  publisher = {Elsevier BV},
}

@InProceedings{Ballini2025,
  author     = {Ballini, E. and Cominelli, A. and Dovera, L. and Forello, A. and Formaggia, L. and Fumagalli, A. and Nardean, S. and Scotti, A. and Zunino, P.},
  booktitle  = {SPE Reservoir Simulation Conference},
  title      = {Enhancing Computational Efficiency of Numerical Simulation for Subsurface Fluid-Induced Deformation Using Deep Learning Reduced Order Models},
  year       = {2025},
  month      = mar,
  publisher  = {SPE},
  series     = {25RSC},
  collection = {25RSC},
  doi        = {10.2118/223897-ms},
}

@Article{Cheng2024,
  author    = {Cheng, Hongrong and Zhang, Miao and Shi, Javen Qinfeng},
  journal   = {IEEE Transactions on Pattern Analysis and Machine Intelligence},
  title     = {A Survey on Deep Neural Network Pruning: Taxonomy, Comparison, Analysis, and Recommendations},
  year      = {2024},
  issn      = {1939-3539},
  month     = dec,
  number    = {12},
  pages     = {10558--10578},
  volume    = {46},
  doi       = {10.1109/tpami.2024.3447085},
  publisher = {Institute of Electrical and Electronics Engineers (IEEE)},
}

@Misc{Gale2019,
  author    = {Gale, Trevor and Elsen, Erich and Hooker, Sara},
  title     = {The State of Sparsity in Deep Neural Networks},
  year      = {2019},
  copyright = {arXiv.org perpetual, non-exclusive license},
  doi       = {10.48550/ARXIV.1902.09574},
  publisher = {arXiv},
}

@Article{Vadera2022,
  author    = {Vadera, Sunil and Ameen, Salem},
  journal   = {IEEE Access},
  title     = {Methods for Pruning Deep Neural Networks},
  year      = {2022},
  issn      = {2169-3536},
  pages     = {63280--63300},
  volume    = {10},
  doi       = {10.1109/access.2022.3182659},
  publisher = {Institute of Electrical and Electronics Engineers (IEEE)},
}

@Misc{Blalock2020,
  author    = {Blalock, Davis and Ortiz, Jose Javier Gonzalez and Frankle, Jonathan and Guttag, John},
  title     = {What is the State of Neural Network Pruning?},
  year      = {2020},
  copyright = {arXiv.org perpetual, non-exclusive license},
  doi       = {10.48550/ARXIV.2003.03033},
  publisher = {arXiv},
}

@PhdThesis{Ballini2025phd,
  author  = {Ballini, Enrico},
  school  = {Politecnico di Milano},
  title   = {Flow and mechanics in fractured porous media: from high fidelity models to efficient reduced order solutions},
  year    = {2025},
  address = {Milan, Italy},
  note    = {Ph.D. Thesis, MOX, Department of Mathematics, Politecnico di Milano, Piazza Leonardo da Vinci 32, 20133 Milano, Italy},
  type    = {Ph.D. thesis},
  url     = {https://hdl.handle.net/10589/232512},
}

@Article{Alqahtani2021,
  author    = {Alqahtani, Ali and Xie, Xianghua and Jones, Mark W.},
  journal   = {Informatics},
  title     = {Literature Review of Deep Network Compression},
  year      = {2021},
  issn      = {2227-9709},
  month     = nov,
  number    = {4},
  pages     = {77},
  volume    = {8},
  doi       = {10.3390/informatics8040077},
  publisher = {MDPI AG},
}

@InProceedings{Mauch2017,
  author    = {Mauch, Lukas and Yang, Bin},
  booktitle = {2017 IEEE International Conference on Acoustics, Speech and Signal Processing (ICASSP)},
  title     = {A novel layerwise pruning method for model reduction of fully connected deep neural networks},
  year      = {2017},
  month     = mar,
  pages     = {2382--2386},
  publisher = {IEEE},
  doi       = {10.1109/icassp.2017.7952583},
}

@Article{Reed1993,
  author    = {Reed, R.},
  journal   = {IEEE Transactions on Neural Networks},
  title     = {Pruning algorithms-a survey},
  year      = {1993},
  issn      = {1045-9227},
  number    = {5},
  pages     = {740--747},
  volume    = {4},
  doi       = {10.1109/72.248452},
  publisher = {Institute of Electrical and Electronics Engineers (IEEE)},
}

@InProceedings{Xiao2019,
  author    = {Xiao, Xia and Wang, Zigeng and Rajasekaran, Sanguthevar},
  booktitle = {Advances in Neural Information Processing Systems},
  title     = {AutoPrune: Automatic Network Pruning by Regularizing Auxiliary Parameters},
  year      = {2019},
  editor    = {H. Wallach and H. Larochelle and A. Beygelzimer and F. d\textquotesingle Alch\'{e}-Buc and E. Fox and R. Garnett},
  publisher = {Curran Associates, Inc.},
  volume    = {32},
  url       = {https://proceedings.neurips.cc/paper_files/paper/2019/file/4efc9e02abdab6b6166251918570a307-Paper.pdf},
}

@InProceedings{Hassibi1992,
  author    = {Hassibi, Babak and Stork, David},
  booktitle = {Advances in Neural Information Processing Systems},
  title     = {Second order derivatives for network pruning: Optimal Brain Surgeon},
  year      = {1992},
  editor    = {S. Hanson and J. Cowan and C. Giles},
  publisher = {Morgan-Kaufmann},
  volume    = {5},
  url       = {https://proceedings.neurips.cc/paper_files/paper/1992/file/303ed4c69846ab36c2904d3ba8573050-Paper.pdf},
}

@InProceedings{Engelbrecht1996,
  author     = {Engelbrecht, A.P. and Cloete, I.},
  booktitle  = {Proceedings of International Conference on Neural Networks (ICNN’96)},
  title      = {A sensitivity analysis algorithm for pruning feedforward neural networks},
  year       = {1996},
  pages      = {1274--1278},
  publisher  = {IEEE},
  series     = {ICNN-96},
  volume     = {2},
  collection = {ICNN-96},
  doi        = {10.1109/icnn.1996.549081},
}

@Article{Zurada1997,
  author    = {Zurada, Jacek M. and Malinowski, Aleksander and Usui, Shiro},
  journal   = {Neurocomputing},
  title     = {Perturbation method for deleting redundant inputs of perceptron networks},
  year      = {1997},
  issn      = {0925-2312},
  month     = feb,
  number    = {2},
  pages     = {177--193},
  volume    = {14},
  doi       = {10.1016/s0925-2312(96)00031-8},
  publisher = {Elsevier BV},
}

@Article{Engelbrecht2001,
  author    = {Engelbrecht, A.P.},
  journal   = {IEEE Transactions on Neural Networks},
  title     = {A new pruning heuristic based on variance analysis of sensitivity information},
  year      = {2001},
  issn      = {1045-9227},
  number    = {6},
  pages     = {1386--1399},
  volume    = {12},
  doi       = {10.1109/72.963775},
  publisher = {Institute of Electrical and Electronics Engineers (IEEE)},
}

@InProceedings{Fnaiech2004,
  author    = {Fnaiech, N. and Abid, S. and Fnaiech, F. and Cheriet, M.},
  booktitle = {First International Symposium on Control, Communications and Signal Processing, 2004.},
  title     = {A modified version of a formal pruning algorithm based on local relative variance analysis},
  year      = {2004},
  pages     = {849--852},
  publisher = {IEEE},
  doi       = {10.1109/isccsp.2004.1296579},
}

@Misc{Babaeizadeh2016,
  author    = {Babaeizadeh, Mohammad and Smaragdis, Paris and Campbell, Roy H.},
  title     = {NoiseOut: A Simple Way to Prune Neural Networks},
  year      = {2016},
  copyright = {arXiv.org perpetual, non-exclusive license},
  doi       = {10.48550/ARXIV.1611.06211},
  info      = {arxiv:1611.06211},
  publisher = {arXiv},
}

@InProceedings{Mauch2018,
  author    = {Mauch, Lukas and Yang, Bin},
  booktitle = {2018 IEEE Statistical Signal Processing Workshop (SSP)},
  title     = {Least-Squares Based Layerwise Pruning Of Convolutional Neural Networks},
  year      = {2018},
  month     = jun,
  pages     = {60--64},
  publisher = {IEEE},
  doi       = {10.1109/ssp.2018.8450814},
}

@InProceedings{Molchanov2019,
  author    = {Molchanov, Pavlo and Mallya, Arun and Tyree, Stephen and Frosio, Iuri and Kautz, Jan},
  booktitle = {2019 IEEE/CVF Conference on Computer Vision and Pattern Recognition (CVPR)},
  title     = {Importance Estimation for Neural Network Pruning},
  year      = {2019},
  month     = jun,
  publisher = {IEEE},
  doi       = {10.1109/cvpr.2019.01152},
}

@Article{Otter2021,
  author    = {Otter, Daniel W. and Medina, Julian R. and Kalita, Jugal K.},
  journal   = {IEEE Transactions on Neural Networks and Learning Systems},
  title     = {A Survey of the Usages of Deep Learning for Natural Language Processing},
  year      = {2021},
  issn      = {2162-2388},
  month     = feb,
  number    = {2},
  pages     = {604--624},
  volume    = {32},
  doi       = {10.1109/tnnls.2020.2979670},
  publisher = {Institute of Electrical and Electronics Engineers (IEEE)},
}

@Article{Voulodimos2018,
  author    = {Voulodimos, Athanasios and Doulamis, Nikolaos and Doulamis, Anastasios and Protopapadakis, Eftychios},
  journal   = {Computational Intelligence and Neuroscience},
  title     = {Deep Learning for Computer Vision: A Brief Review},
  year      = {2018},
  issn      = {1687-5273},
  pages     = {1--13},
  volume    = {2018},
  doi       = {10.1155/2018/7068349},
  publisher = {Wiley},
}

@Article{EdDyyany2025,
  author    = {Ed Dyyany, Ayoub and Jamea, Ahmed and Ammar, Abdelghali},
  journal   = {SHS Web of Conferences},
  title     = {Neural networks for solving partial differential equations, a comprehensive review of recent methods and applications},
  year      = {2025},
  issn      = {2261-2424},
  pages     = {01005},
  volume    = {214},
  doi       = {10.1051/shsconf/202521401005},
  editor    = {Cherradi, B. and Jamea, A. and Boukhair, A. and Deschepper, C.},
  publisher = {EDP Sciences},
}

@InProceedings{Sanh2020,
  author    = {Sanh, Victor and Thomas, Wolf and Rush, Alexander M.},
  booktitle = {Neural Information Processing Systems (NeurIPS)},
  title     = {Movement pruning: Adaptive sparsity by fine-tuning},
  year      = {2020},
}

@Misc{Qian2021,
  author    = {Qian, Xin and Klabjan, Diego},
  title     = {A Probabilistic Approach to Neural Network Pruning},
  year      = {2021},
  copyright = {Creative Commons Attribution 4.0 International},
  doi       = {10.48550/ARXIV.2105.10065},
  notes     = {arXiv:2105.10065},
  publisher = {arXiv},
}

@InProceedings{Mozer1988,
  author    = {Michael C. Mozer and Paul Smolensky},
  booktitle = {Proceedings of the 2nd International Conference on Neural Information Processing Systems (NIPS)},
  title     = {Skeletonization: A Technique for Trimming the Fat from a Network via Relevance Assessment},
  year      = {1988},
  pages     = {107--115},
  doi       = {10.5555/2969735.2969748},
}

@InProceedings{Yu2018,
  author    = {Yu, Ruichi and Li, Ang and Chen, Chun-Fu and Lai, Jui-Hsin and Morariu, Vlad I. and Han, Xintong and Gao, Mingfei and Lin, Ching-Yung and Davis, Larry S.},
  booktitle = {2018 IEEE/CVF Conference on Computer Vision and Pattern Recognition},
  title     = {NISP: Pruning Networks Using Neuron Importance Score Propagation},
  year      = {2018},
  month     = jun,
  publisher = {IEEE},
  doi       = {10.1109/cvpr.2018.00958},
}

@Article{Janowsky1989,
  author    = {Janowsky, Steven A.},
  journal   = {Physical Review A},
  title     = {Pruning versus clipping in neural networks},
  year      = {1989},
  issn      = {0556-2791},
  month     = jun,
  number    = {12},
  pages     = {6600--6603},
  volume    = {39},
  doi       = {10.1103/physreva.39.6600},
  publisher = {American Physical Society (APS)},
}

@TechReport{LeCun1998,
  author      = {LeCun, Yann and Bottou, L{\'e}on and Bengio, Yoshua and Haffner, Patrick},
  institution = {AT\&T Labs},
  title       = {Gradient-Based Learning Applied to Document Recognition},
  year        = {1998},
  number      = {CRP-TR-98-11},
}

@Misc{Takamoto2022,
  author    = {Takamoto, Makoto and Praditia, Timothy and Leiteritz, Raphael and MacKinlay, Dan and Alesiani, Francesco and Pflüger, Dirk and Niepert, Mathias},
  title     = {PDEBENCH: An Extensive Benchmark for Scientific Machine Learning},
  year      = {2022},
  copyright = {arXiv.org perpetual, non-exclusive license},
  doi       = {10.48550/ARXIV.2210.07182},
  journal   = {arXiv:2210.07182},
  publisher = {arXiv},
}

@Misc{mnistweb,
  author       = {Yann LeCun and Corinna Cortes and Christopher J.~C. Burges},
  howpublished = {\url{https://yann.lecun.org/exdb/mnist/}},
  note         = {Accessed: 2025-12-18},
  title        = {MNIST Handwritten Digit Database},
  year         = {2010},
}

@Misc{Karlbauer2021,
  author    = {Karlbauer, Matthias and Praditia, Timothy and Otte, Sebastian and Oladyshkin, Sergey and Nowak, Wolfgang and Butz, Martin V.},
  title     = {Composing Partial Differential Equations with Physics-Aware Neural Networks},
  year      = {2021},
  copyright = {Creative Commons Attribution Non Commercial No Derivatives 4.0 International},
  doi       = {10.48550/ARXIV.2111.11798},
  publisher = {arXiv},
}

\end{document}